\documentclass{article}

\usepackage[preprint]{neurips_2025}

\usepackage[utf8]{inputenc} 
\usepackage[T1]{fontenc}    
\usepackage{hyperref}       
\usepackage{url}            
\usepackage{booktabs}       
\usepackage{amsfonts}       
\usepackage{graphicx}       
\usepackage{nicefrac}       
\usepackage{microtype}      
\usepackage{xcolor}         
\usepackage{amsmath}        
\usepackage{mathrsfs}       

\usepackage[most]{tcolorbox}     
\tcbuselibrary{skins, breakable}

\usepackage{multirow}       
\usepackage{threeparttable} 

\title{LLMs for Supply Chain Management}

\author{%
  Haojie Wang\\
  School of Management\\
  Harbin Institute of Technology\\
  \texttt{23b910008@stu.hit.edu.cn} \\
  \And
    Jiuyun Jiang\\
  School of Management\\
  Harbin Institute of Technology\\
  \texttt{jiuyunjiang@stu.hit.edu.cn} \\
  \And  
  L. Jeff Hong\\
  Department of Industrial and \\ 
  Systems Engineering\\
  University of Minnesota\\
  \texttt{lhong@umn.edu} \\
  \And  
  Guangxin Jiang\\\
  School of Management\\
  Harbin Institute of Technology\\
  \texttt{gxjiang@hit.edu.cn} \\
}

\begin{document}

\maketitle

\begin{abstract}
 The development of large language models (LLMs) has provided new tools for research in supply chain management (SCM). In this paper, we introduce a retrieval-augmented generation (RAG) framework that dynamically integrates external knowledge into the inference process, and develop a domain-specialized SCM LLM, which demonstrates expert-level competence by passing standardized SCM examinations and beer game tests. We further employ the use of LLMs to conduct horizontal and vertical supply chain games, in order to analyze competition and cooperation within supply chains. Our experiments show that RAG significantly improves performance on SCM tasks. Moreover, game-theoretic analysis reveals that the LLM can reproduce insights from the classical SCM literature, while also uncovering novel behaviors and offering fresh perspectives on phenomena such as the bullwhip effect. This paper opens the door for exploring cooperation and competition for complex supply chain network through the lens of LLMs.
\end{abstract}

\section{Introduction}
Supply chain management (SCM) is at the core of business operations. With increasing uncertainty in global economic development and the deepening of digital transformation, supply chain networks have become increasingly large and diverse in structure, encompassing multiple stages from raw material procurement to product delivery. Enterprises must not only address demand uncertainty, supply risks, and information asymmetry, but also coordinate the interests of multiple tiers and agents to achieve efficient, flexible, and sustainable operations. Moreover, data fragmentation, issues of timeliness and accuracy, and the ever-changing market and policy environment further exacerbate the difficulties of supply chain decision-making \citep{davis2023replication, balabanova2025media}. 

With the rapid advancement of LLMs, their remarkable capabilities in knowledge integration, information retrieval, complex reasoning, and multi-agent communication have opened new avenues for tackling complex decision-making challenges in SCM \citep{duan2024gtbench, li2024empowering}. In particular, retrieval-augmented generation (RAG) techniques can further enhance LLM performance by incorporating external knowledge during inference. 
However, the systematic application of LLMs to SCM tasks, especially those involving multilevel, multi-agent interactions and game-theoretic mechanisms, remains underexplored. In the context of vertical (hierarchical) supply chain games, research is especially limited, particularly in settings characterized by information asymmetry, heterogeneous risk preferences, and dynamic feedback mechanisms. Furthermore, the utility boundaries and limitations of mainstream LLMs in such vertical scenarios have not been thoroughly examined, highlighting the need for more intelligent, context-aware decision-making approaches in SCM.

To address these gaps, we develop a domain-specialized SCM LLM based on a RAG framework. The model is trained and evaluated to exhibit expert-level competence, as evidenced by its ability to pass standardized SCM exams and beer game tests. To the best of our knowledge, this is the first LLM designed for the supply chain management domain. Then, we employ the LLM to simulate both horizontal and vertical supply chain games, enabling the analysis of strategic behavior related to competition and cooperation. Through multi-agent experiments, we evaluate model performance across diverse scenarios encompassing varied information structures and risk preferences, and assess their effectiveness in tasks such as behavioral experiment replication and multi-level interaction simulation. 

\paragraph{Main Contributions}
Our main contributions are as follows:
\begin{itemize}
    \item We develop the first domain-specialized SCM LLM, trained and evaluated to demonstrate expert-level competence by passing standardized exams.
    \item We propose a new paradigm for studying complex decision-making in SCM using LLMs, with a particular focus on competition and cooperation.
    \item We conduct comprehensive experiments to investigate the capabilities and limitations of LLMs in SCM.
\end{itemize}

\section{Related Works}

Our work advances the application of LLMs in SCM, focusing on RAG, horizontal and vertical game theory. We review related literature as follows.

RAG was proposed to address LLMs’ limitations in knowledge coverage and factuality~\citep{borgeaud2022improving, izacard2021distilling, khandelwal2019generalization, izacard2021distilling, shi2024replug}. Despite advances in language understanding, generation, recommender systems~\citep{zhao2024recommender, wang2024rethinking}, and molecular discovery~\citep{li2024empowering}, LLMs still face issues such as domain-specific knowledge scarcity, hallucinations~\citep{dahl2024large, zhao2024recommender, zhao2023survey}, and inefficient updates. RAG mitigates these problems by retrieving and integrating external knowledge during generation, improving performance in question answering~\citep{borgeaud2022improving, izacard2021leveraging, liu2022uni}, dialogue~\citep{chen2020bridging, ghazvininejad2018knowledge, komeili2022internet, zhang2020grounded}, fact verification~\citep{asai2023self, izacard2023atlas, lewis2020retrieval}, and recommender systems~\citep{di2023retrieval, lu2021revcore, wu2024coral}, while reducing hallucinations~\citep{shuster2021retrieval, xu2022beyond}. Applications of RAG now span NLP, AI for Science~\citep{li2024empowering, liu2023multi, lozano2023clinfo, wang2023retrieval, wang2024biobridge}, finance~\citep{li2024alphafin, lin2024revolutionizing, yepes2024financial, zhang2023enhancing}, and medicine~\citep{wu2024medical}. Yet, its use in supply chain scenarios remains limited due to the lack of high-quality, domain-specific knowledge bases. To address this, our work constructs a dedicated supply chain knowledge base to promote RAG’s practical application in this field.

Game theory offers a systematic framework for modeling strategic interactions in supply chains~\citep{balabanova2025media, falcao2024making, han2022emergent}. The integration of LLMs into game-theoretic settings has enabled new studies on cooperation, negotiation, and competition among agents~\citep{duan2024gtbench, xie2024can, xu2024magic}, with Nash equilibrium~\citep{nash1950equilibrium, risse2000rational} providing a basis for evaluating LLMs’ decisions. However, most existing work still focuses on simple, classical games—such as the Prisoner's Dilemma and Ultimatum Game—limiting practical relevance and raising concerns about test set leakage~\citep{guo2023gpt, phelps2023investigating, akata2025playing, aher2023using, brookins2024playing}. More recent research has started to introduce complex, dynamic multi-agent games~\citep{huang2025competing}, but connecting theory with reproducible empirical studies remains a challenge~\citep{mao2023alympics}. Distinct from prior benchmarks, our study leverages DeepSeek-R1 to systematically explore horizontal game-theoretic interactions in supply chains and applies reinforcement learning to enhance LLM-based decision-making.

In SCM, horizontal game analysis is well-established, typically focusing on Nash equilibrium concepts. In contrast, vertical games are less studied and mostly limited to replicating behavioral experiments. Applications of LLMs in this area are also scarce. Recent studies have replicated large-scale behavioral experiments in supply chains, inventory management, and queueing systems, confirming the robustness of behavioral effects~\citep{davis2023replication}. Multi-agent LLM systems (e.g., InvAgent) further leverage zero-shot reasoning to simulate vertical interactions in multi-echelon inventory systems, though they still focus on classical phenomena such as the bullwhip effect~\citep{quan2024invagent}.

In summary, LLM-based vertical supply chain game analysis remains limited to behavioral replication and bullwhip effect studies, with systematic exploration of more complex vertical mechanisms still lacking. Our research aims to expand LLM applications in vertical supply chain analysis by validating classical bullwhip effect experiments~\citep{croson2006behavioral} and further exploring the impact of risk preferences and information completeness. 

\section{Supply Chain Vertical LLMs}

To adapt LLMs to the highly specialized field of SCM, it is essential to systematically integrate domain expertise, advanced retrieval mechanisms, and rigorous evaluation methods. In this section, we outline the construction of a supply chain knowledge base, the implementation of an RAG framework, and the design of a professional capability assessment—laying the groundwork for vertical LLMs in SCM. All the details are provided in Appendix A.

\subsection{Domain Expertise Database}
SCM is a highly specialized field, and constructing a professional knowledge base is fundamental for building vertical LLMs. To ensure authoritative and comprehensive coverage, we integrated unstructured domain knowledge primarily from textbooks and training materials for internationally recognized qualifications such as SCMP and CPIM, systematically encompassing core modules like supply chain planning, execution, inventory control, and logistics management. Chinese and English encyclopedia knowledge was also incorporated to broaden coverage. Using semantic vectorization, we construct the above unstructured domain knowledge into a vector database, which facilitated subsequent retrieval and utilization.

\subsection{Retrieval-Augmented Generation}
Building on the supply chain expertise base, we leverage RAG technology to adapt leading open-source LLMs to this domain and enhance their capabilities. Our hybrid retrieval strategy combines keyword matching with semantic similarity retrieval to improve the precision of knowledge acquisition. For factual queries, we prioritize exact matching, while for decision-making queries, we emphasize semantic relevance. Retrieval results are integrated and deduplicated through a context fusion layer, then injected into open-source LLMs such as LLaMa 3.1 and DeepSeek using structured prompts. A dynamic context length adjustment mechanism adaptively tunes the number of retrieved documents based on query complexity, ensuring optimal resource utilization and maintaining high-quality reasoning, thereby enabling efficient performance of the supply chain vertical LLM in specialized tasks.

\subsection{Professional Capability Tests}
In order to verify the actual professional capabilities of supply chain vertical LLMs, we design a two-dimensional evaluation system. In terms of professional knowledge testing, we select real questions from SCM certification exams such as SCMP and CPIM over the years as the test set. Table~\ref{tab1} shows that after adopting RAG technology, the accuracy of the model in professional knowledge testing has been significantly improved\footnote{Requirement for passing SCMP exam is 60\%, and requirement for passing CPIM exam is 75\%.}. In particular, supply chain vertical LLMs using RAG technology have better performance in issues such as the interpretation of professional terms and the selection of best practices. 

\begin{table}[htbp]
\centering
\caption{Performance (\%) of Different LLMs on SCMP and CPIM with/without RAG} 
\label{tab1}
\begin{threeparttable}
\begin{tabular}{lcc}
\toprule
\textbf{Model} & \textbf{SCMP} & \textbf{CPIM} \\
\midrule
LLaMa-3.1-8b (w/o RAG) & 28.57\% & 46.67\% \\
LLaMa-3.1-8b (w/ RAG)  & \bf 44.29\% & \bf 60.83\% \\
DeepSeek-R1-70b (w/o RAG) & 55.71\% & 68.33\% \\
DeepSeek-R1-70b (w/ RAG)  & \bf 77.86\% & \bf 81.67\% \\
\bottomrule
\end{tabular}
\end{threeparttable}
\end{table}

\begin{figure}[h]
  \centering
  \includegraphics[width=10cm]{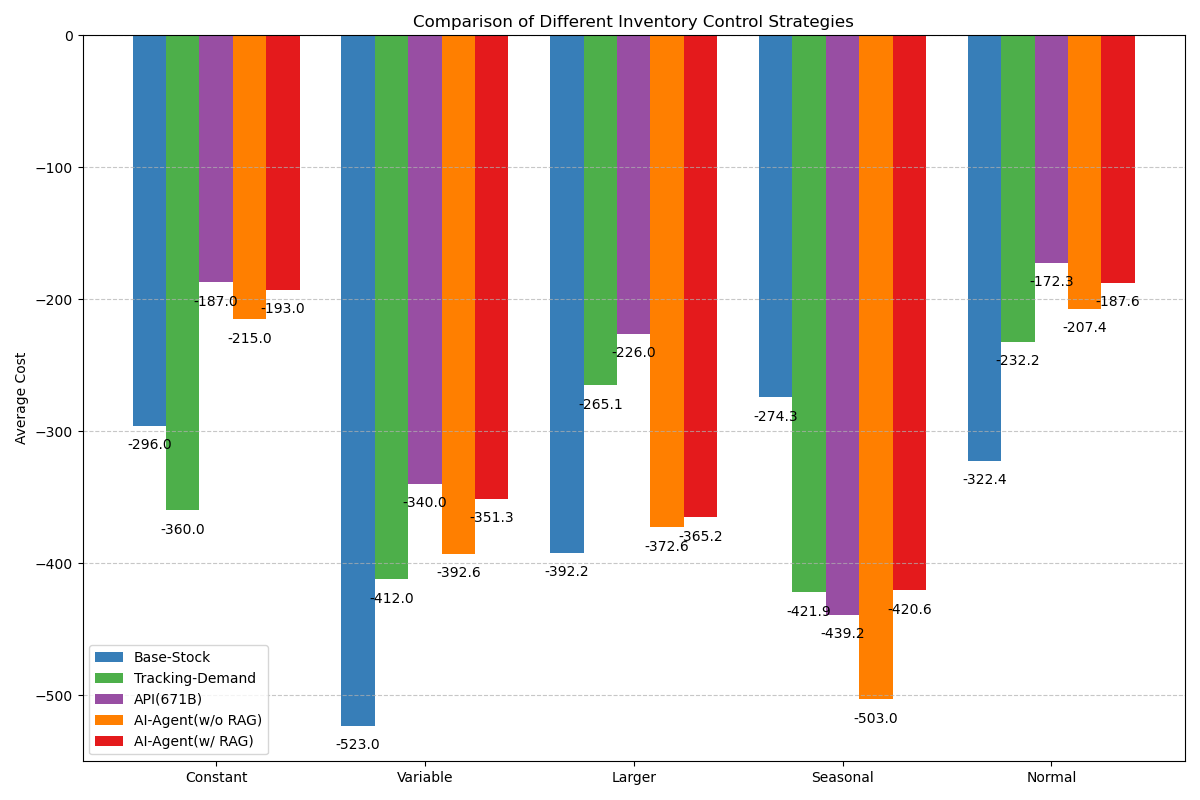}
  \caption{Management professional ability performance.}
  \label{fig1}
\end{figure}

The decision-making ability test is based on the classic beer game. We construct a variety of market demand scenarios, including typical supply chain challenges such as stable demand, range demand, seasonal fluctuations, etc. In addition, we compare the decisions made by LLMs with those derived from classical strategies such as base-stock and tracking-demand policies. For more details about the beer game and experimental settings, refer to Appendix A.3. Figure~\ref{fig1} shows the decision-making potential of LLMs in the face of these challenges. Our supply chain vertical LLMs using RAG perform better in most scenarios.

\section{Horizontal and Vertical Games}



Supply chain interactions can be horizontal (between firms at the same stage) or vertical (across different tiers). Analyzing both types allows us to effectively evaluate the decision-making and interaction capabilities of LLMs in complex supply chain scenarios. In this section, we use horizontal competition and vertical dynamics to examine how effectively LLMs can navigate and optimize multi-agent supply chain scenarios. To accelerate the convergence of the LLM, we restrict its numerical outputs to integer values.

\subsection{Horizontal Game: Competition among Multiple Suppliers}

Horizontal games in supply chains examine competition among firms at the same stage, typically in quantity or price. Modeling these dynamics is essential for capturing real-world market behavior. Two classic frameworks—Cournot and Bertrand—are commonly used, with the Cournot model focusing on quantity-based competition among suppliers. The experimental details are provided in Appendix B.

\subsubsection{Cournot Model in Supply Chain}

The Cournot model describes a market structure in which a limited number of firms (oligopolists) simultaneously decide the quantity of output to produce, while taking into account the total market supply and the anticipated behavior of competitors. Each firm aims to maximize its own profit, assuming the output of its rivals is fixed. The interaction leads to a Nash equilibrium, where no firm can unilaterally increase its profit by altering its output decision. This model is especially relevant for analyzing competition in supply chains where firms offer homogeneous products and have limited market information.

\paragraph{Experiment design}
With the assistance of a LLM, we designed and conducted an output game experiment to simulate horizontal competition among suppliers at a given supply chain tier. Market price is determined by a linear inverse demand function ($P = a - bQ$), where $Q$ is the total market output and market capacity is set at 100 units. Two scenarios are considered: \textbf{Duopoly}, in which two suppliers with zero production cost ($cost=0$) independently decide their outputs over 10 rounds without knowledge of each other's decisions; and \textbf{Oligopoly with heterogeneous costs}, where three suppliers with different production costs ($cost = [15, 20, 25]$) compete under the same information constraints, introducing cost asymmetry to the game dynamics. In both scenarios, supplier agents are powered by a LLM, enabling adaptive learning and strategic adjustment based on historical observations.

\begin{figure}[h]
	\centering
	\begin{minipage}[b]{0.45\textwidth}
		\centering
		\includegraphics[width=\textwidth]{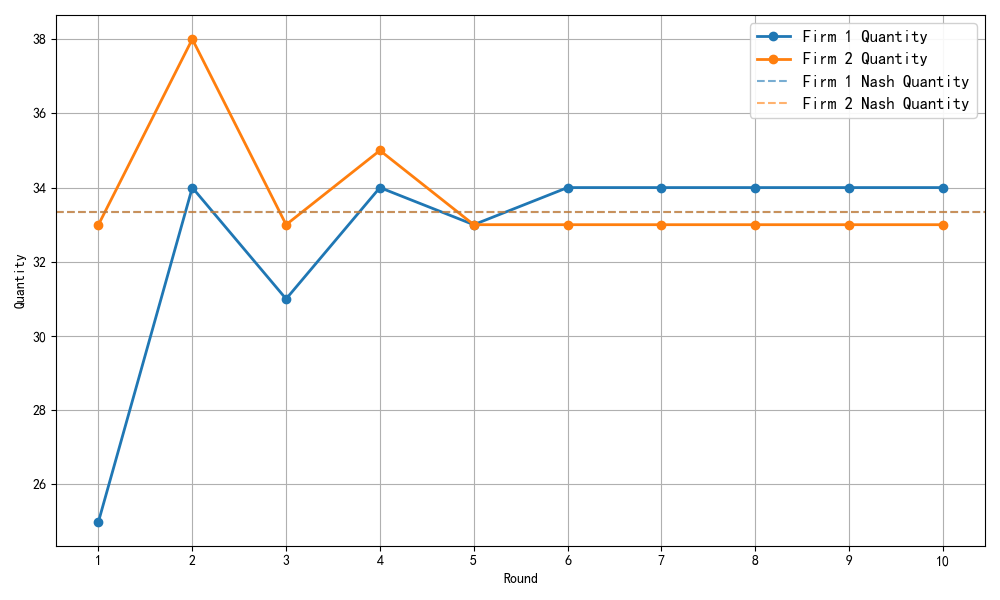}
		\vspace{1pt}
		\small (a) Production Decision
	\end{minipage}
	\hfill
	\begin{minipage}[b]{0.45\textwidth}
		\centering
		\includegraphics[width=\textwidth]{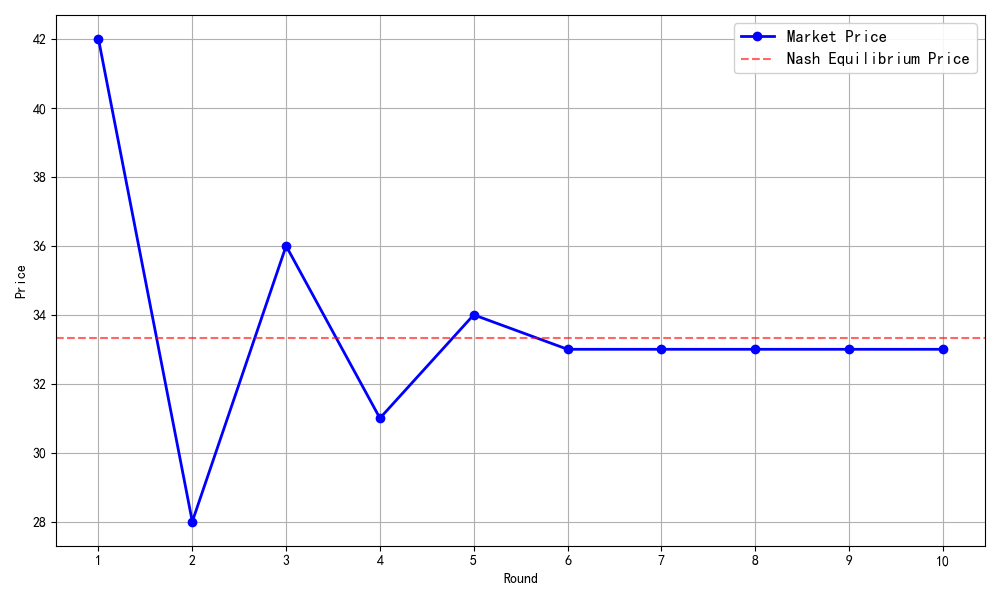}
		\vspace{1pt}
		\small (b) Market Price
	\end{minipage}
	\caption{Supplier game under the duopoly Cournot model.}
	\label{fig2}
\end{figure}

\begin{figure}[h]
	\centering
	\begin{minipage}[b]{0.45\textwidth}
		\centering
		\includegraphics[width=\textwidth]{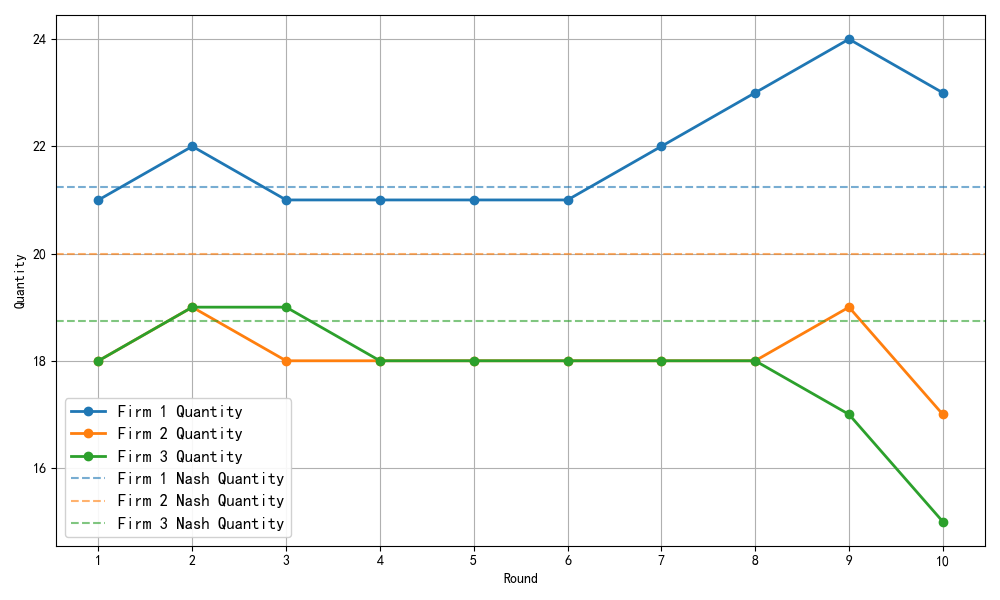}
		\vspace{1pt}
		\small (a) Production Decision
	\end{minipage}
	\hfill
	\begin{minipage}[b]{0.45\textwidth}
		\centering
		\includegraphics[width=\textwidth]{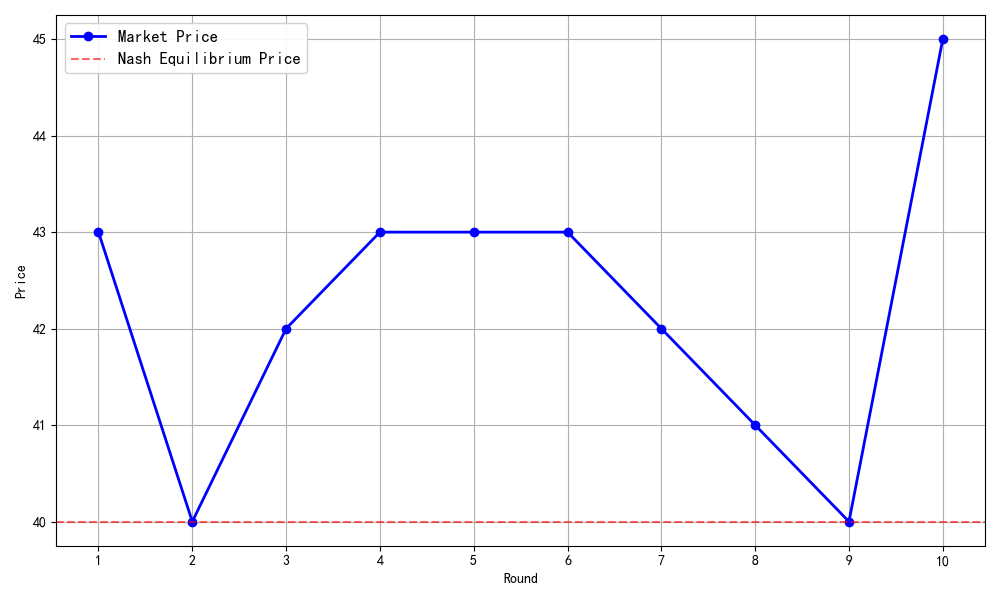}
		\vspace{1pt}
		\small (b) Market Price
	\end{minipage}
	\caption{Supplier game under Cournot model of multiple oligopolies ($n=3$).}
	\label{fig3}
\end{figure}

\paragraph{Results analysis}
The results for the duopoly scenario are shown in Figure~\ref{fig2}, and the results for the oligopoly scenario are shown in Figure~\ref{fig3}. It can be seen that after several rounds of competition, the output decisions of the companies in both scenarios eventually tend to the Cournot-Nash equilibrium, as predicted by theory. Although the output decisions are more volatile due to increased complexity and information asymmetry in the case of three suppliers with heterogeneous costs, the output still tends to fluctuate around their respective Nash equilibrium values. As shown in Figure~\ref{fig2}(b) and Figure~\ref{fig3}(b), the market price also fluctuates, but eventually tends to the theoretical equilibrium.

\subsubsection{Bertrand Model in Supply Chain}

The Bertrand model is a fundamental framework for analyzing horizontal price competition among firms at the same level of a supply chain. In this model, a limited number of suppliers simultaneously decide on product prices rather than quantities, under the assumption of homogeneous or differentiated products and rational consumer behavior. Consumers always choose the supplier offering the lower price. The Bertrand model is particularly relevant for markets where price wars are prevalent and marginal costs play a critical role in determining competitive advantage.

\paragraph{Experiment design}
With the assistance of a LLM, we simulate a series of pricing competition experiments between two oligopolistic suppliers in a supply chain context. Market demand follows a linear function ($Q = a - bP$) with a fixed market size of 100 units. Two scenarios are considered: \textbf{cost asymmetry}, where two suppliers produce homogeneous products but have different production costs ($cost = [20, 40]$), setting prices independently over 10 rounds without access to each other's decisions; and \textbf{product heterogeneity}, where both suppliers have identical costs ($cost = [20, 20]$) but their products differ in substitutability, represented by heterogeneity coefficients ($heterogeneity~coefficient = [0.5, 1.5]$), with higher values indicating less substitutable products. In both scenarios, pricing and sales volumes are recorded over 10 rounds, while all other market conditions are held constant to ensure comparability of results.

\begin{figure}[h]
	\centering
	\begin{minipage}[b]{0.45\textwidth}
		\centering
		\includegraphics[width=\textwidth]{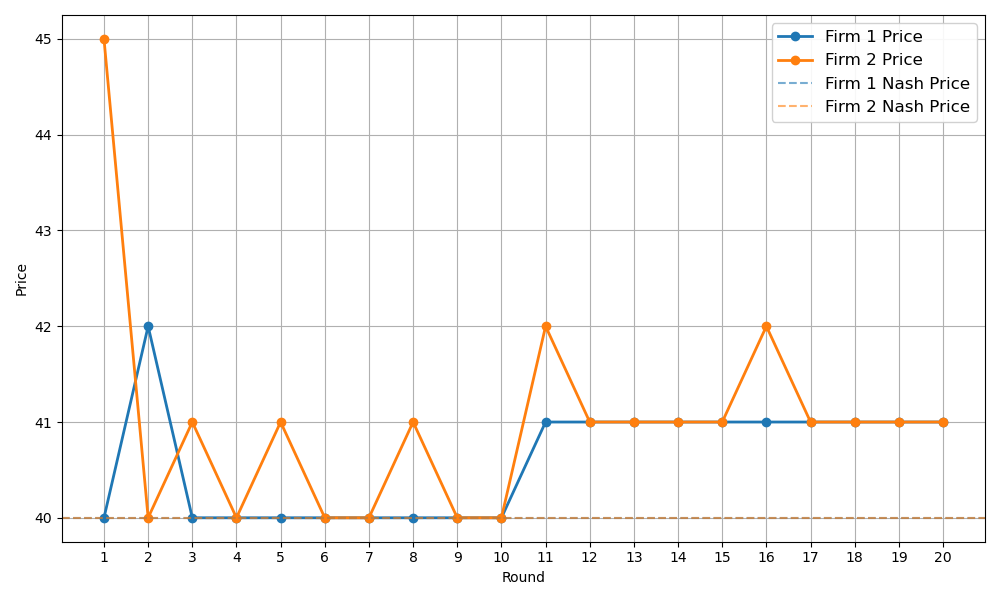}
		\vspace{1pt}
		\small (a) Price Decision
	\end{minipage}
	\hfill
	\begin{minipage}[b]{0.45\textwidth}
		\centering
		\includegraphics[width=\textwidth]{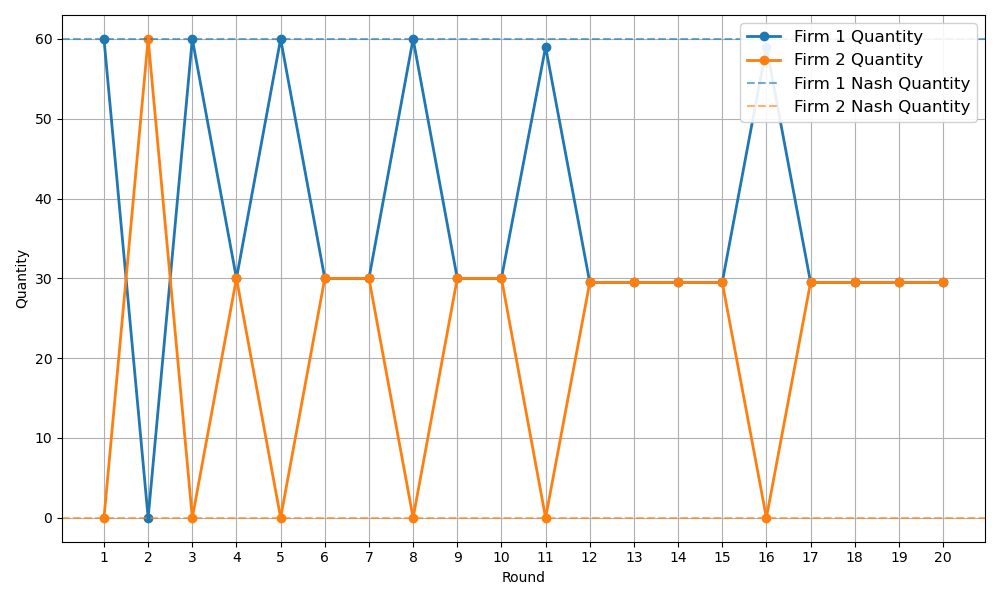}
		\vspace{1pt}
		\small (b) Sales Volume Changes
	\end{minipage}
	\caption{Supplier game under Bertrand model.}
	\label{fig4}
\end{figure}

\paragraph{Results analysis}
In the cost asymmetry scenario, as shown in Figure~\ref{fig4}, both suppliers' prices initially fluctuate as they adapt to the competitive environment. Over several rounds, the prices converge towards the Nash equilibrium values. The lower-cost supplier (Firm 1) attains a dominant position, leveraging its cost advantage to either capture the entire market or set higher prices for greater profit, depending on the rival’s strategy. This results in the sales volumes oscillating between the "winner-takes-all" equilibrium and the unstable state of equal market division.

\begin{figure}[h]
	\centering
	\begin{minipage}[b]{0.45\textwidth}
		\centering
		\includegraphics[width=\textwidth]{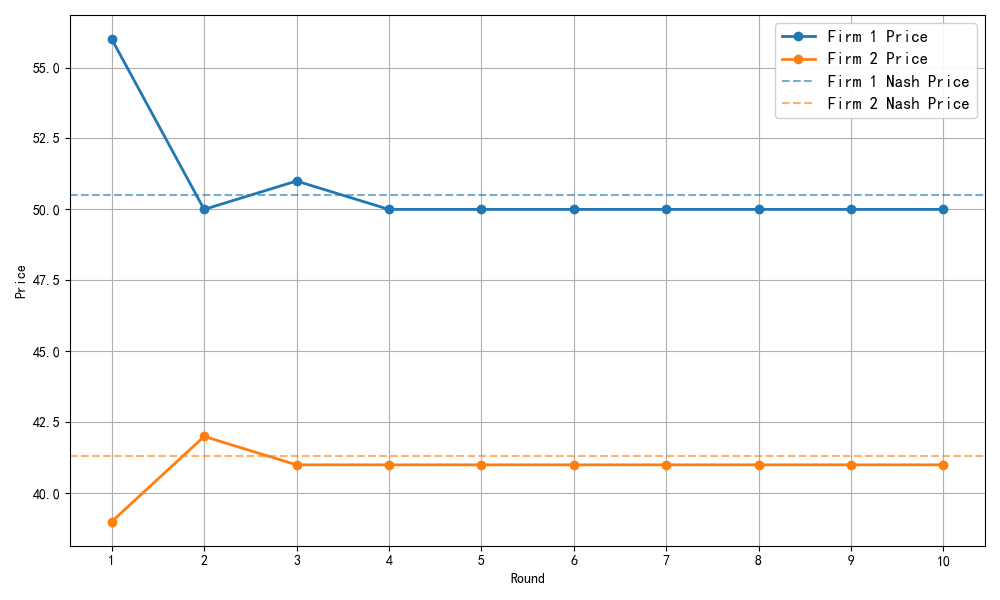}
		\vspace{1pt}
		\small (a) Price Decision
	\end{minipage}
	\hfill
	\begin{minipage}[b]{0.45\textwidth}
		\centering
		\includegraphics[width=\textwidth]{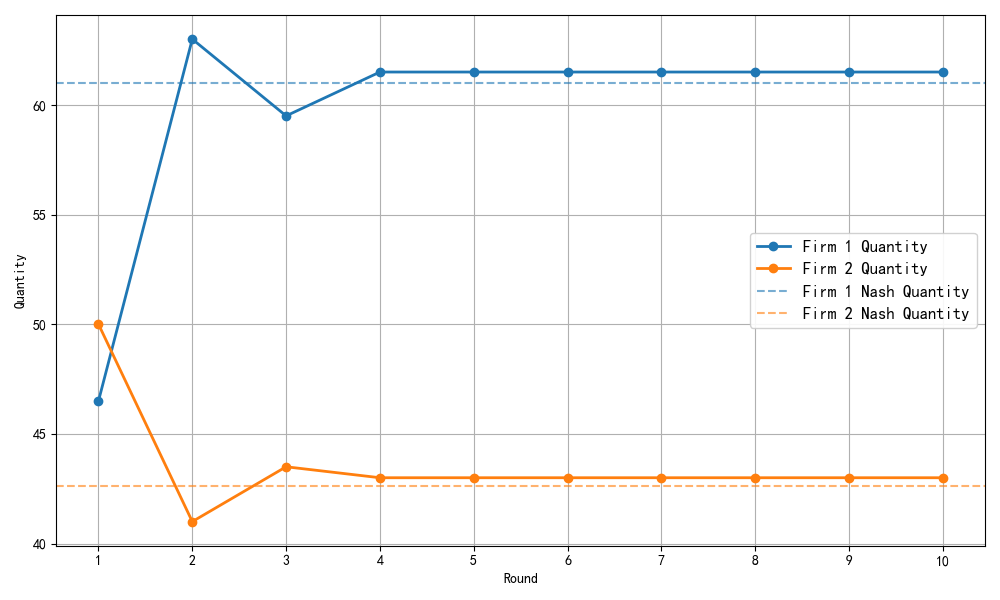}
		\vspace{1pt}
		\small (b) Sales Volume Changes
	\end{minipage}
	\caption{Supplier game with asymmetric heterogeneity coefficient under Bertrand model.}
	\label{fig5}
\end{figure}

In the scenario involving product heterogeneity, depicted in Figure~\ref{fig5}, both suppliers have identical costs but different heterogeneity coefficients. The supplier with the higher heterogeneity coefficient (greater product irreplaceability) is able to sustain a higher price and achieves a larger market share. After 10 rounds, both pricing and sales volume stabilize around the Nash equilibrium, demonstrating the significant impact of product differentiation on market power and equilibrium outcomes.

\subsection{Vertical Game: Bullwhip Effect}

The bullwhip effect is a well-known phenomenon in SCM, where demand fluctuations are amplified as information flows upstream in the supply chain. Suppliers and manufacturers often face greater order variability than the actual customer demand at the retail level. This can lead to issues like inventory buildup, production inefficiencies, and resource waste, driven by factors such as demand forecast errors, order batching, price fluctuations, and information asymmetry.

In this study, we adopt the initial experimental design from the relevant 2006 literature by employing the beer game to test for the existence of the bullwhip effect \citep{croson2006behavioral}. However, unlike previous studies that focused solely on behavioral experiments and replications of the original setup, our main contribution lies in introducing a vertical large-scale supply chain model to systematically test its utility boundaries. Specifically, we examine the model’s performance under different risk preferences and levels of information sharing as the follows. The experimental details are provided in Appendix C.

\subsubsection{Bullwhip Effect under Different Risk Preferences with Isolated Information}

\paragraph{Experiment design}
To investigate the bullwhip effect under different risk preferences and isolated information, we design a classical supply chain experiment based on the beer distribution game. The supply chain comprises four stages: retailer, wholesaler, distributor, and manufacturer. Over 24 periods, the retailer faces market demand from a uniform distribution $U[0,8]$ and orders upstream, while only the retailer knows the demand distribution; upstream stages decide with limited information. Stage lead times are $[2,2,2,3]$, initial inventory is 12 units, and production capacity is 20 units per period for each stage. Holding and backlog costs are set at 0.5 and 1 per unit per period, respectively. We consider three agent types: \textbf{Risk aversion}, prioritizing stock safety and over-ordering to avoid stockouts; \textbf{Risk neutrality}, balancing inventory and backlog costs for optimal efficiency; and \textbf{Risk appetite}, minimizing inventory while accepting higher risks of stockouts and backlog penalties.

\paragraph{Results analysis}

\begin{figure}[h]
	\centering
	\begin{minipage}[b]{0.45\textwidth}
		\centering
		\includegraphics[width=\textwidth]{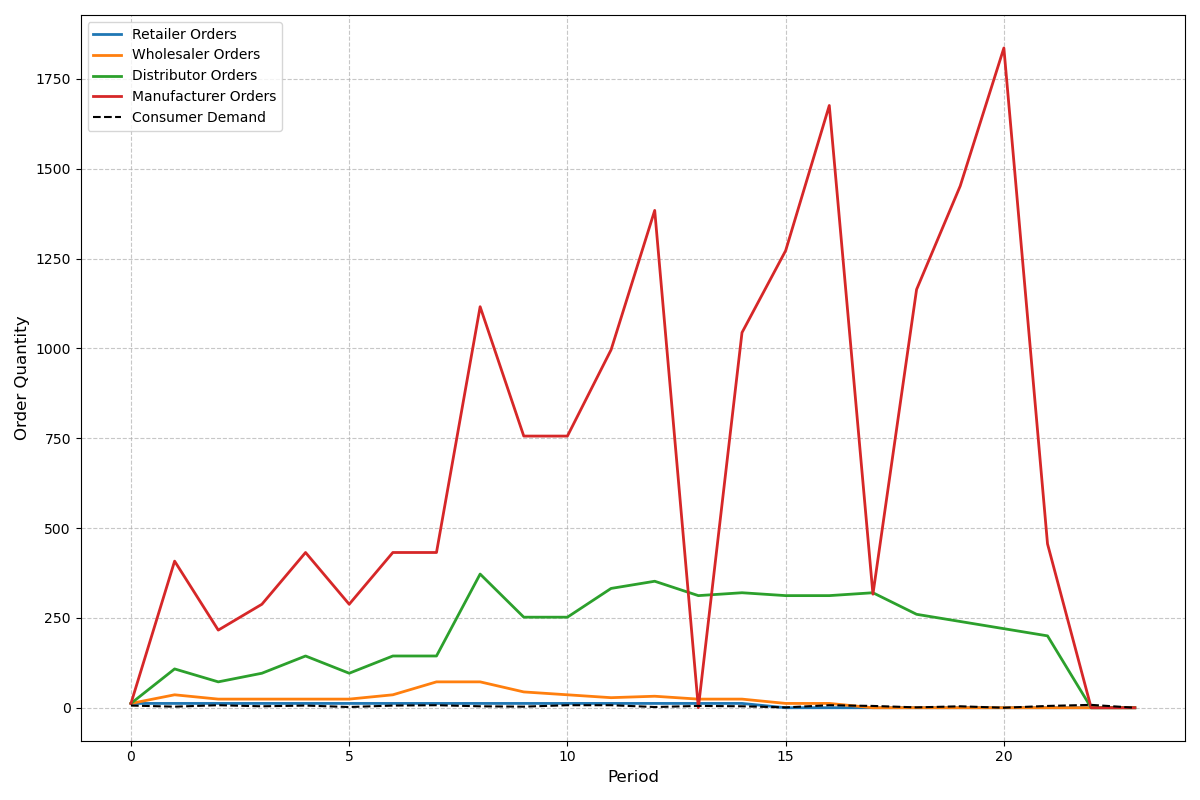}
		\vspace{1pt}
		\small (a) Risk Aversion
	\end{minipage}
	\hfill
	\begin{minipage}[b]{0.45\textwidth}
		\centering
		\includegraphics[width=\textwidth]{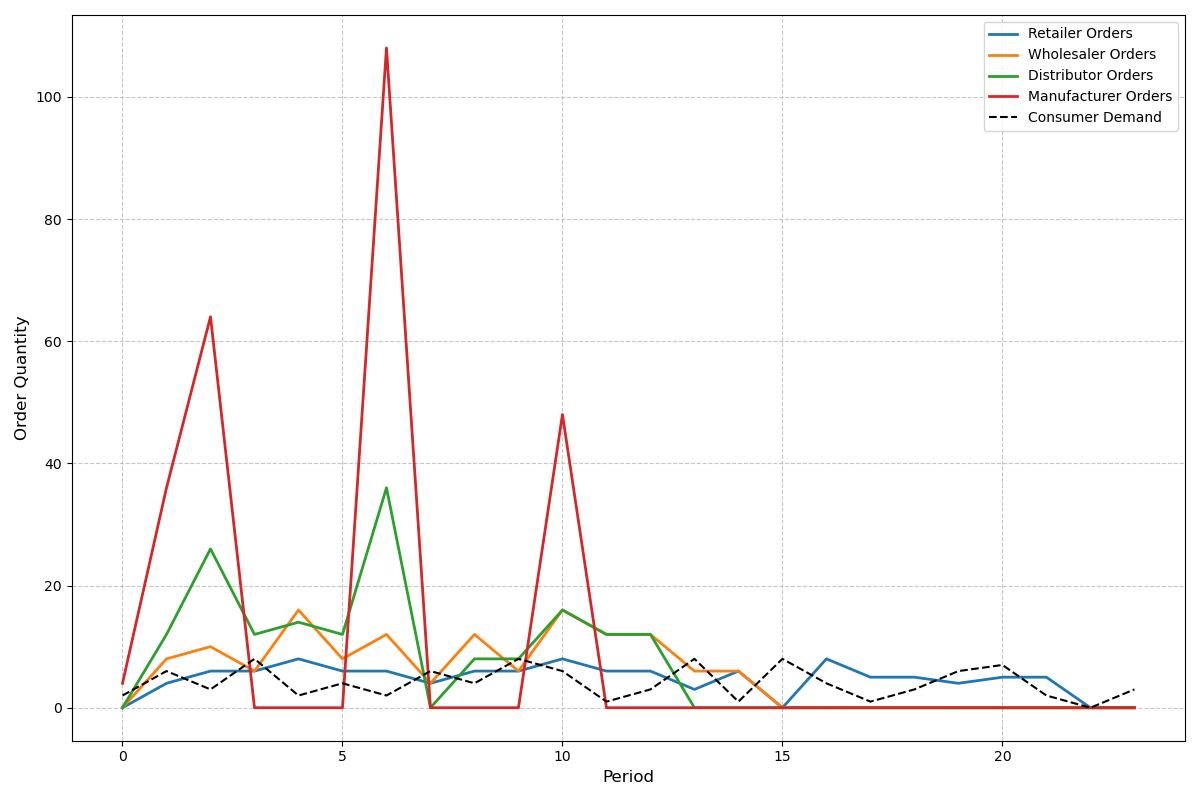}
		\vspace{1pt}
		\small (b) Risk Neutral
	\end{minipage}
	\hfill
	\begin{minipage}[b]{0.45\textwidth}
		\centering
		\includegraphics[width=\textwidth]{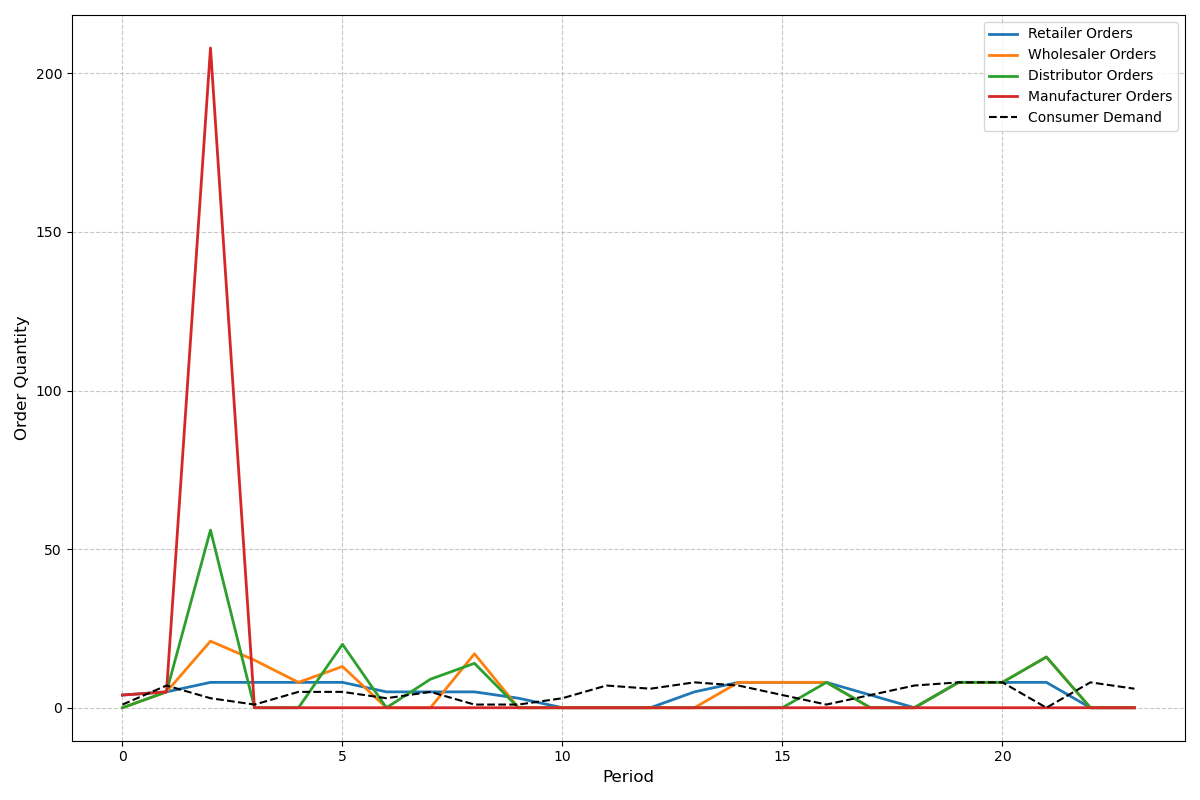}
		\vspace{1pt}
		\small (c) Risk Appetite
	\end{minipage}
	\hfill
	\begin{minipage}[b]{0.45\textwidth}
		\centering
		\includegraphics[width=\textwidth]{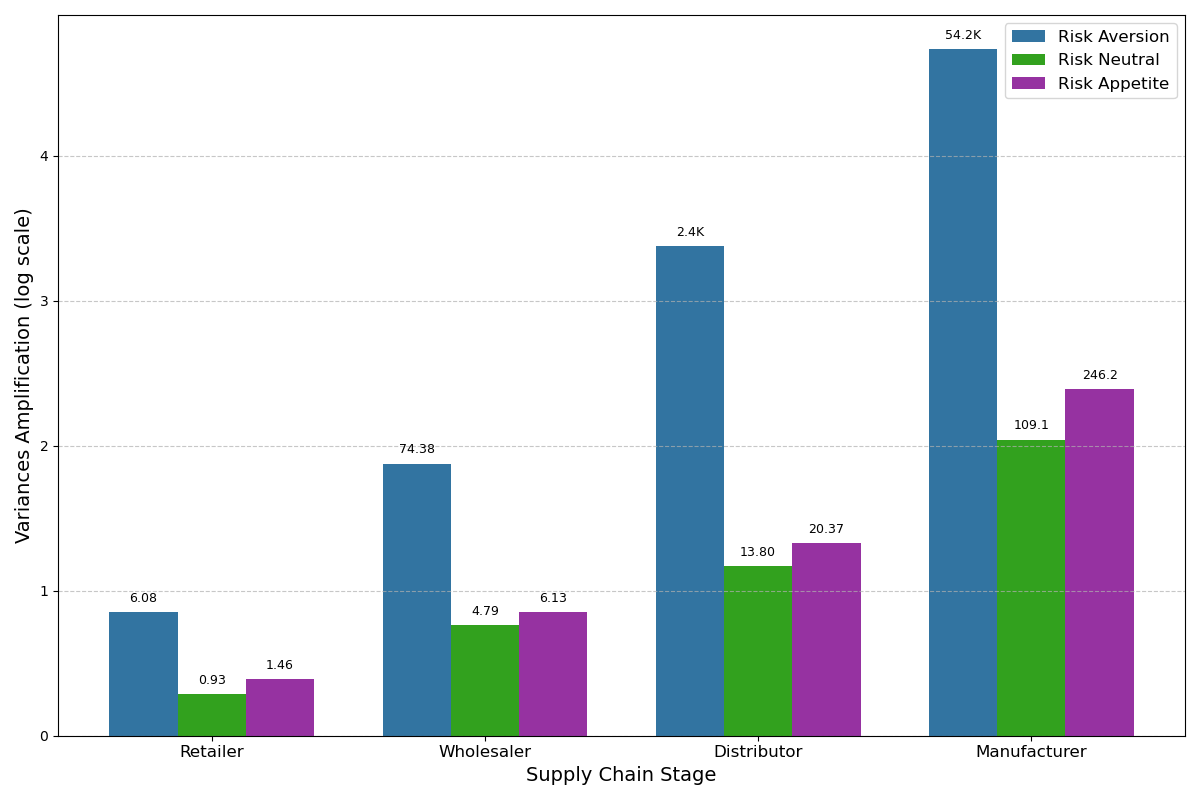}
		\vspace{1pt}
		\small (d) Variances Amplification
	\end{minipage}
	\caption{Bullwhip effect under different risk preferences with isolated information.}
	\label{fig6}
\end{figure}

The experimental results, as shown in Figure~\ref{fig6}, compare the bullwhip effect across the three risk preference scenarios.

In the risk-averse case (Figure~\ref{fig6}(a)), the bullwhip effect is most pronounced, with dramatic amplification and volatility in upstream order quantities—particularly at the manufacturer stage—due to excessive safety stock ordering. This behavior amplifies demand fluctuations throughout the supply chain, leading to significant inventory accumulation and pronounced order spikes. In contrast, under risk neutrality (Figure~\ref{fig6}(b)), the bullwhip effect is substantially mitigated; order quantities at all stages remain relatively stable with minimal variance amplification, reflecting a balanced trade-off between inventory safety and holding costs and resulting in the lowest bullwhip intensity among the three scenarios. For risk-appetite agents (Figure~\ref{fig6}(c)), lower inventories and less aggressive ordering reduce holding costs but cause sporadic, sharp replenishment when inventories are depleted, leading to periodic surges in upstream orders. Accordingly, the bullwhip effect is moderate—greater than in the risk-neutral case, but less severe than under risk aversion.

Figure~\ref{fig6}(d) quantifies variance amplification at each supply chain stage under different risk preferences, clearly showing that risk aversion leads to the highest amplification, particularly upstream, while risk neutrality yields the most stable supply chain response. Risk-appetite agents display intermediate amplification. Overall, these results demonstrate that risk preference is a key determinant of the bullwhip effect when information is isolated: risk-averse behavior exacerbates order variability and inventory accumulation, whereas risk-neutral strategies are most effective in stabilizing supply chain dynamics.

\subsubsection{Bullwhip Effect under Different Risk Preferences with Information Sharing}

\paragraph{Experiment design}
To evaluate the impact of information sharing on the bullwhip effect under varying risk preferences, we extend the classic four-stage beer distribution supply chain experiment by granting each stage access to real-time information regarding the inventory, backlog, and lead times of its immediate upstream partner. This enhanced transparency enables each agent to make more informed ordering decisions by considering upstream conditions. All other experimental parameters remain consistent with the baseline scenario involving isolated information.

\begin{figure}[h]
	\centering
	\begin{minipage}[b]{0.45\textwidth}
		\centering
		\includegraphics[width=\textwidth]{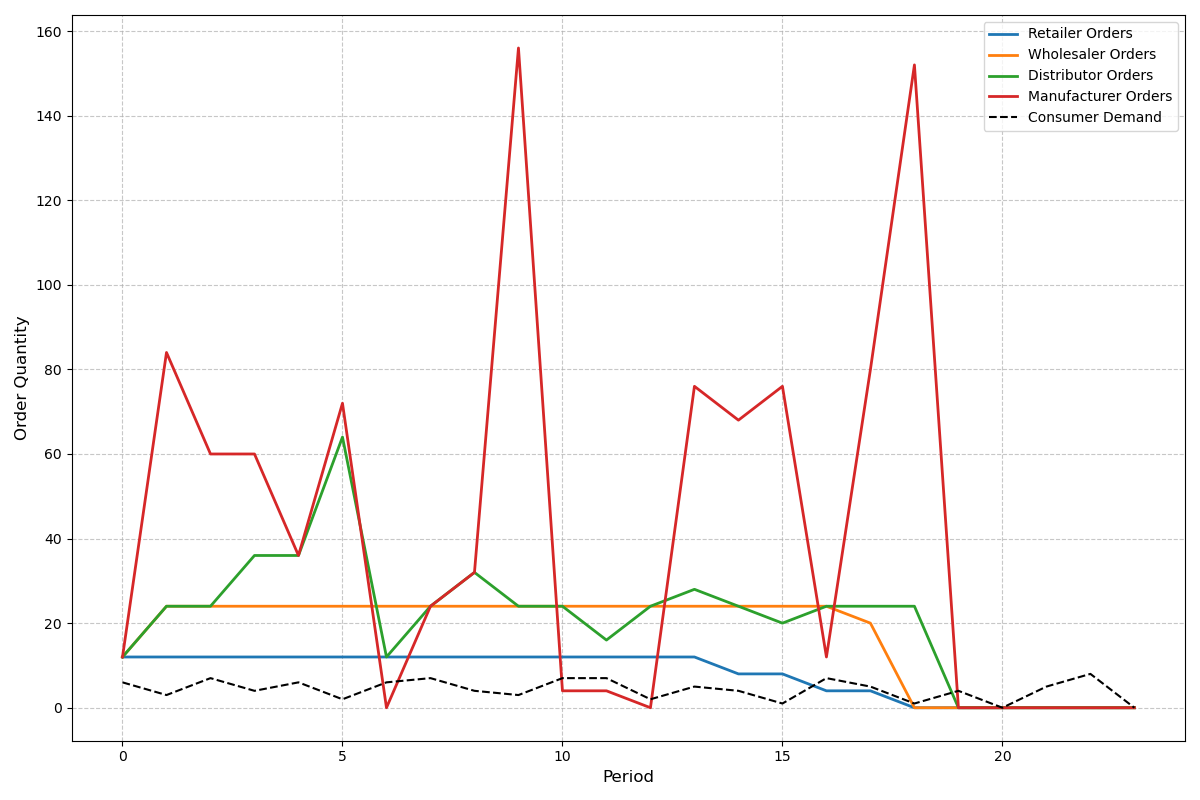}
		\vspace{1pt}
		
		\small (a) Risk Aversion
	\end{minipage}
	\hfill
	\begin{minipage}[b]{0.45\textwidth}
		\centering
		\includegraphics[width=\textwidth]{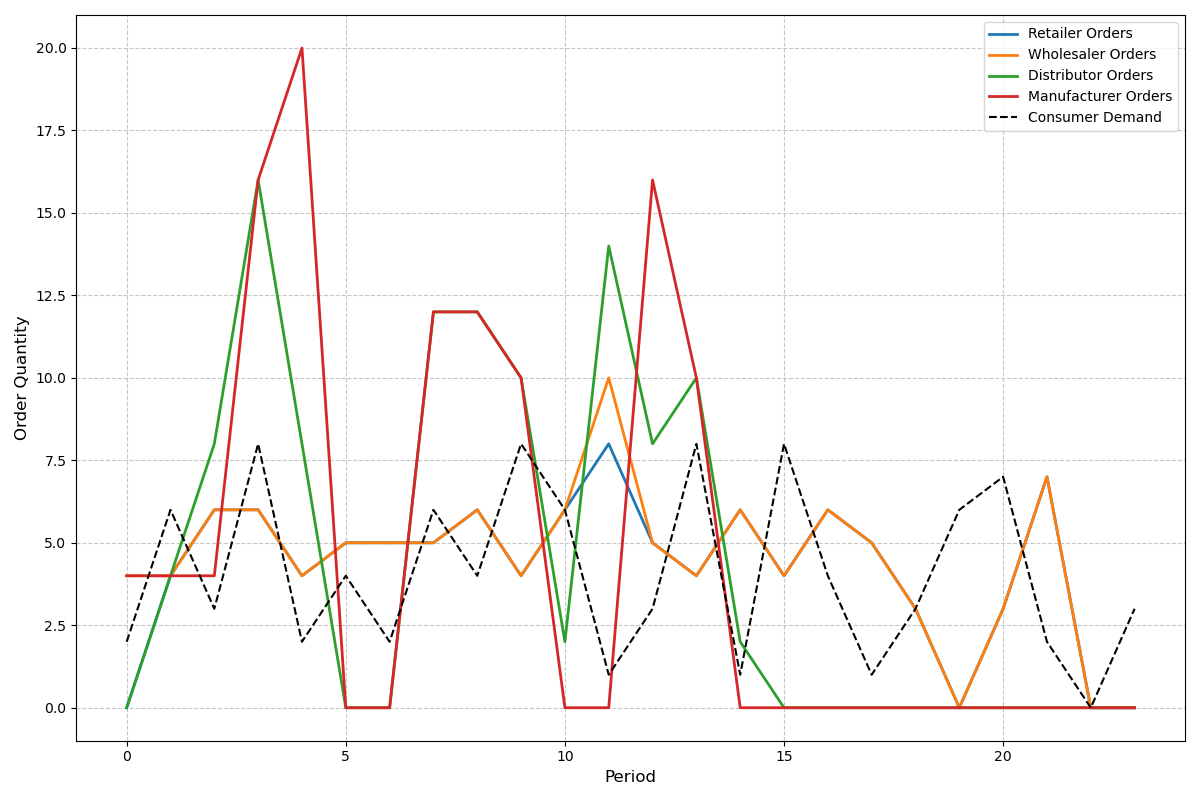}
		\vspace{1pt}
		
		\small (b) Risk Neutral
	\end{minipage}
	\hfill
	\begin{minipage}[b]{0.45\textwidth}
		\centering
		\includegraphics[width=\textwidth]{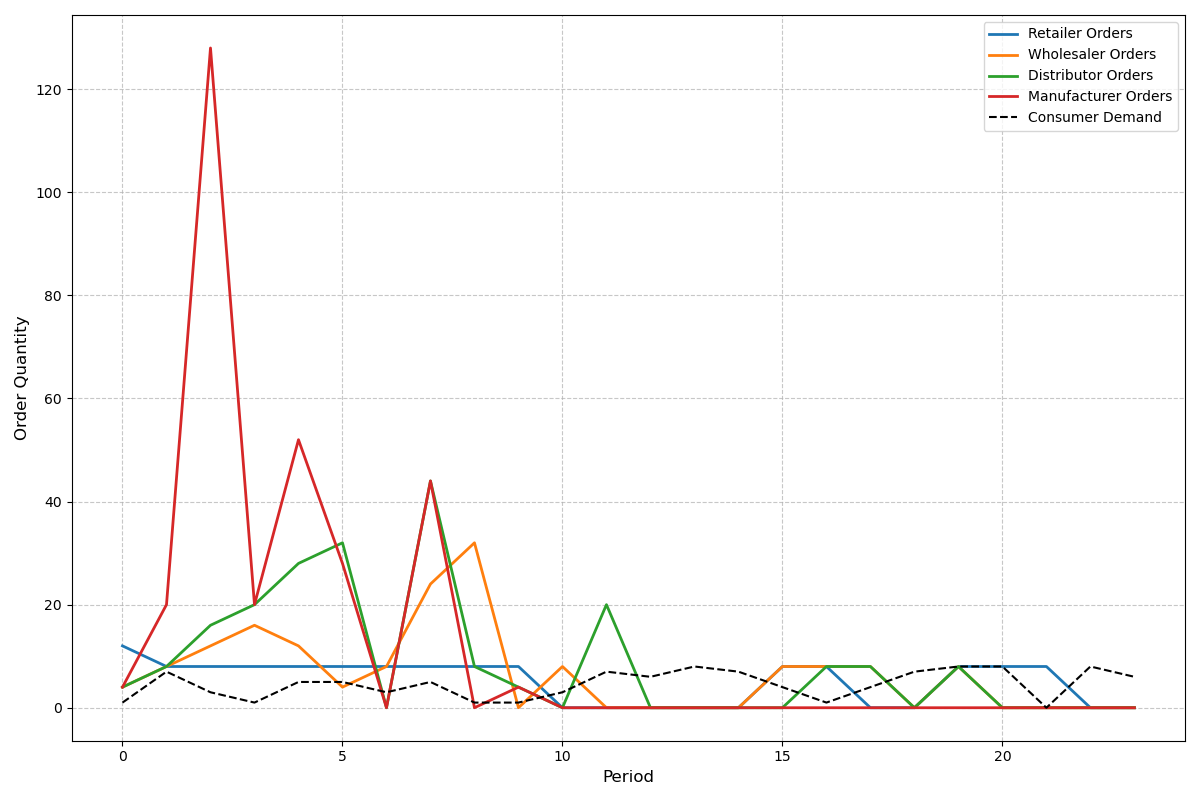}
		\vspace{1pt}
		\small (c) Risk Appetite
	\end{minipage}
	\hfill
	\begin{minipage}[b]{0.45\textwidth}
		\centering
		\includegraphics[width=\textwidth]{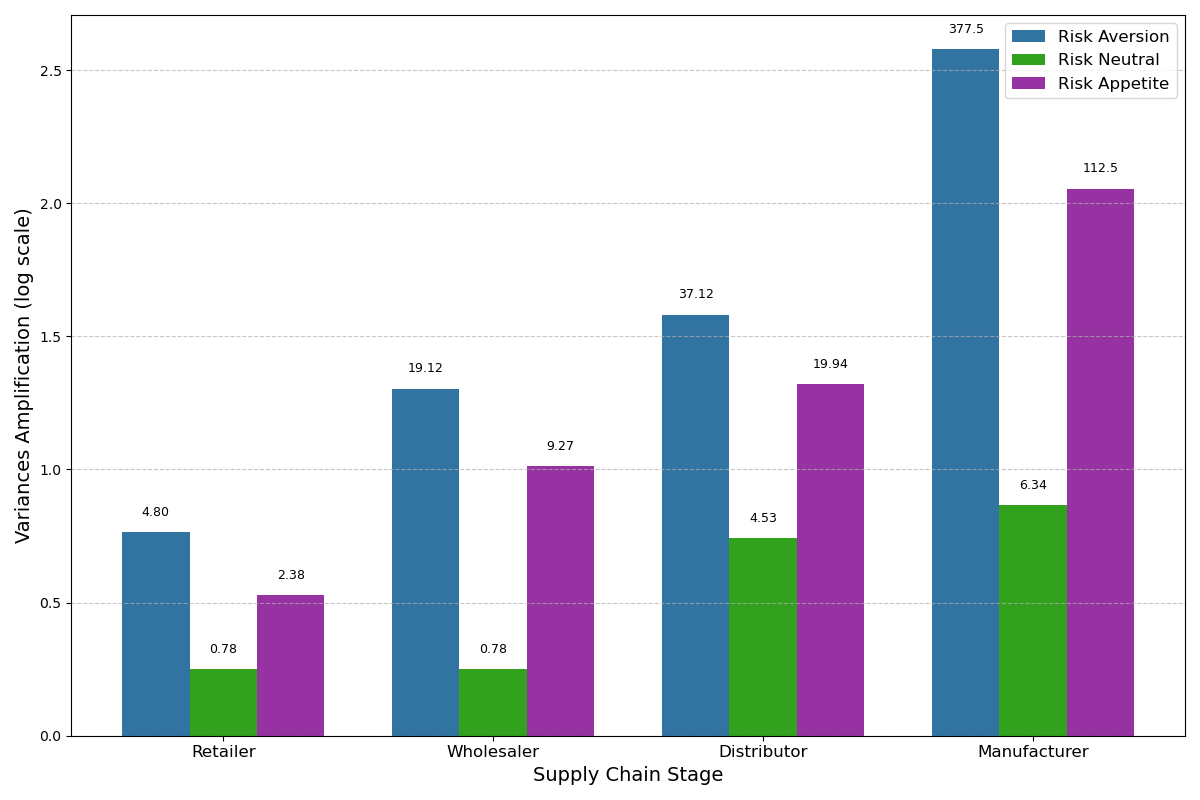}
		\vspace{1pt}
		\small (d) Variances Amplification
	\end{minipage}
	\caption{Bullwhip effect under different risk preferences with information sharing.}
	\label{fig7}
\end{figure}

\paragraph{Results analysis}
Figure~\ref{fig7} presents the bullwhip effect results under information sharing for the three risk preference types.

With information sharing, the bullwhip effect remains for risk-averse agents (Figure~\ref{fig7}(a)), but is significantly reduced compared to scenarios without shared information. Upstream order amplification and variability decrease as agents coordinate more effectively and rely less on excessive safety stock, thanks to greater visibility of inventory and backlog. For risk-neutral agents (Figure~\ref{fig7}(b)), order quantities become highly stable and closely track end-customer demand, minimizing variance and demonstrating that information sharing with balanced risk preferences produces the most stable supply chain. For risk-seeking agents (Figure~\ref{fig7}(c)), although occasional large replenishment orders still occur, overall order variability declines noticeably; improved transparency enables better anticipation of supply chain needs and restrains extreme or erratic ordering behavior.

Figure~\ref{fig7}(d) quantitatively compares variance amplification across all risk preferences, illustrating that information sharing significantly reduces the bullwhip effect at every stage. Risk-neutral agents show the least amplification, while risk-averse agents remain more susceptible, though the impact is much less severe than without information sharing. Overall, these results confirm that information sharing is highly effective in mitigating the bullwhip effect across different risk profiles, enabling supply chains to achieve greater stability and efficiency through transparency and rational decision-making.

\begin{table}[htbp]
	\centering
	\caption{Summary of Bullwhip Effect Variance Amplification under Different Risk Preferences and Information Conditions}
	\label{tab2}
	\begin{tabular}{cccccc}
		\toprule
		\textbf{Risk Preference} & \textbf{Stage} & \textbf{w/o IS} & \textbf{w/ IS} & \textbf{Information Gain} \\
		\midrule
		\multirow{4}{*}{Risk Aversion} 
		& Retailer      & 6.08$\times$    & 4.80$\times$    & 21.05\%   \\
		& Wholesaler    & 74.38$\times$   & 19.12$\times$   & 74.29\%   \\
		& Distributor   & 2389.26$\times$ & 37.12$\times$   & 98.45\%   \\
		& Manufacturer  & 54216.12$\times$& 377.52$\times$  & 99.30\%   \\
		\midrule
		\multirow{4}{*}{Risk Neutral}
		& Retailer      & 0.93$\times$    & 0.78$\times$    & 16.13\%   \\
		& Wholesaler    & 4.79$\times$    & 0.78$\times$    & 83.72\%   \\
		& Distributor   & 13.80$\times$   & 4.53$\times$    & 67.17\%   \\
		& Manufacturer  & 109.06$\times$  & 6.34$\times$    & 94.19\%   \\
		\midrule
		\multirow{4}{*}{Risk Appetite}
		& Retailer      & 1.46$\times$    & 2.38$\times$    & $-$63.01\% \\
		& Wholesaler    & 6.13$\times$    & 9.27$\times$    & $-$51.22\% \\
		& Distributor   & 20.37$\times$   & 19.94$\times$   & 2.11\%     \\
		& Manufacturer  & 246.15$\times$  & 112.47$\times$  & 54.31\%    \\
		\bottomrule
	\end{tabular}
\end{table}

\paragraph{Comprehensive Analysis}

Table~\ref{tab2} summarizes variance amplification (the bullwhip effect) across supply chain stages under three risk preference regimes, with and without information sharing. Information sharing markedly reduces the bullwhip effect for risk-averse and risk-neutral agents—especially at the manufacturer stage, where order variance drops by up to 99.3\% and 94.19\%, respectively, consistent with previous findings~\citep{croson2006behavioral}. For risk-seeking agents, however, its impact is less predictable: in some cases, variability may even increase, suggesting that risk appetite can lead to overreactions when more information is available. Still, a significant variance reduction (54.31\%) is observed for risk-seeking agents at the manufacturer stage. Overall, these results indicate that information sharing is most effective for risk-averse and risk-neutral agents, while its benefits for risk-seeking agents are more variable. Tailoring information policies to agent risk profiles may therefore be important for optimal supply chain performance.

These findings also highlight that LLMs are powerful tools in studying SCM problems. By efficiently analyzing complex, multi-stage data and simulating a range of behavioral scenarios, LLMs can identify patterns in variability, anticipate the effects of information policies, and offer tailored recommendations for different agent profiles. This demonstrates the value of LLM-powered analytics in enhancing supply chain stability, optimizing information sharing strategies, and enabling more informed decision-making across diverse risk environments.

\section{Conclusion}



This paper explores the application of LLMs to the specialized domain of SCM, with a focus on integrating domain-specific knowledge and modeling complex multi-agent interactions. We develop the first domain-specialized SCM LLM, grounded in a RAG framework and supported by a curated supply chain knowledge base. The model demonstrates expert-level competence in standardized assessments and exhibits strong performance in reasoning and decision-making tasks. We employ the model to simulate both horizontal and vertical supply chain games, enabling detailed analysis of strategic behavior under varying information structures and risk preferences. Our experiments show that RAG-augmented LLMs not only replicate classical SCM phenomena—such as the bullwhip effect—but also yield novel insights. For example, we find that information sharing mitigates the bullwhip effect among risk-neutral and risk-averse agents, while its impact may diminish or even reverse for risk-seeking agents. In horizontal settings, LLM agents adaptively exhibit cooperative or competitive behavior depending on the incentive structure and information environment, highlighting their versatility in modeling complex supply chain dynamics. These findings suggest that LLMs, when properly adapted, can serve as powerful tools for SCM research and decision support, exemplifying the AI for management science paradigm. Future work should explore the scalability of LLM-based multi-agent systems, expand domain-specific knowledge bases, and design more rigorous evaluation frameworks to bridge the gap between theoretical modeling and practical SCM applications.


\bibliographystyle{unsrtnat}
\bibliography{ref}






\newpage
\appendix
\setcounter{figure}{0}
\setcounter{table}{0}
\section*{Appendix A\quad SCM LLM with RAG}

\paragraph{Background}
Since the advent of large language models (LLMs), people have expected them to have the ability to solve diverse tasks. Recently, LLMs have shown their potential to solve specialized problems in vertical fields. We focus on supply chain management (SCM) because this field requires an extremely broad and interdisciplinary knowledge base: supply chain experts must master logistics, procurement, risk management, finance, and organizational leadership at the same time. Even experienced experts often need to integrate the latest information from a large number of contracts, transaction records, market reports, and policy documents. This is where RAG technology comes into play: by efficiently retrieving relevant information across domains and generating contextual and actionable responses, RAG can help integrate cross-domain knowledge and enhance decision-making capabilities.

\paragraph{RAG System Construction}
RAG is crucial for domains where accuracy and traceability matter. By connecting a specialized retriever with a robust LLM generator, we get the best of both worlds:
\begin{itemize}
    \item \textbf{Relevance:} High-relevance knowledge retrieval.
    \item \textbf{Comprehensibility:} Reasoning-rich response generation.
    \item \textbf{Reliability:} Reduced hallucinations through direct referencing.
\end{itemize}

\begin{figure}[h]
\renewcommand{\thefigure}{\Roman{figure}} 
  \centering
  \includegraphics[width=14cm]{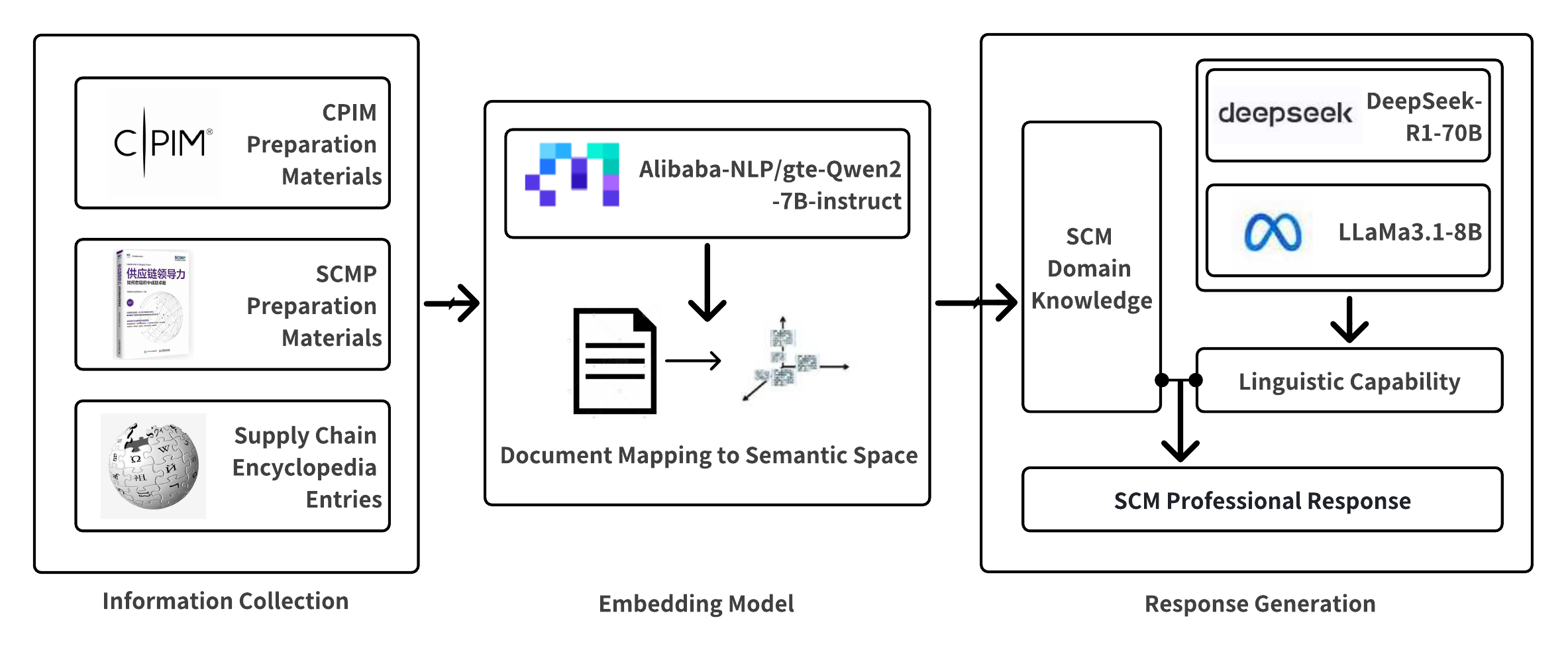}
  \caption{RAG System Diagram.}
  \label{fig_RAG}
\end{figure}

To build RAG System, we performed the tasks (as shown in Figure \ref{fig_RAG}):
\begin{itemize}
    \item \textbf{Information collection:} We use the preparation materials for the SCMP and CPIM exams as a domain knowledge base to ensure the accuracy of the LLM's domain knowledge. SCMP is a professional qualification certification exam established by the China Federation of Logistics and Purchasing (CFLP) for supply chain management experts. It is also the only localized professional certification program in China's supply chain field. SCMP training and examinations cover four core modules: procurement management, production operations, logistics management, and supply chain strategy and planning, as well as multiple sub-fields. This comprehensive knowledge system ensures that certificate holders have extensive general knowledge and in-depth professional capabilities. Specifically, its training and examinations include three compulsory modules: supply chain operations, supply chain planning, and supply chain leadership, and three optional modules: logistics management, planning management, and procurement management. Passing the exam will be awarded the SCMP certificate, which is widely regarded as one of the highest-level certificates in China's supply chain management field. CPIM is a professional qualification certification assessment organized by the Association for Supply Chain Management, the largest non-profit association in the supply chain industry. It is one of the most influential certifications in the field of supply chain management. In order to comprehensively evaluate the candidates' knowledge system, the CPIM exam includes eight modules: supply chains and strategy, sales and operations planning, demand, supply, detailed schedules, inventory, distribution, quality, technology, and continuous improvement. Passing the exam will be awarded the CPIM certificate, which is not only the world's first widely recognized supply chain management certificate, but also the highest-rated supply chain-related certificate in the market. Many corporate supply chain management managers regard the CPIM certificate as the starting point of their careers.  In order to avoid falsely high performance when testing the professional knowledge of SCM LLM, {\em we deliberately deleted the data of historical actual exam questions, simulation exam questions and after-class exercises when sorting out the preparation materials for SCMP and CPIM}. This ensures that RAG can only provide domain knowledge for SCM LLM, but cannot reveal the exam answers. At the same time, in order to ensure the comprehensiveness of domain knowledge, we obtained 2,917 encyclopedia data in the supply chain field through crawler technology.
    \item \textbf{Embedding Model:} We use \textit{Alibaba-NLP/gte-Qwen2-7B-instruct} as the embedding model, which is the current best embedding model on the MTEB leaderboard. Although reasoning LLMs have recently performed surprisingly well on many tasks, their performance on embedding tasks is not optimistic. This may be because reasoning LLMs focus on sequential thinking and logical connections, and therefore cannot map documents into semantic space well. In contrast, \textit{gte-Qwen2-7B-instruct} is specifically trained for the semantic similarity task, creating a high-dimensional space where documents with similar connotations are tightly clustered together, regardless of the specific wording. This means that reasoning LLMs may follow the reasoning path when performing embedding tasks, resulting in content with the same topic but different semantic connotations, resulting in poor performance. On the other hand, \textit{gte-Qwen2-7B-instruct} is able to better capture the intent behind the query. Therefore, we use the latter to complete the embedding operation.
    \item \textbf{Response Generation:} Generate responses using LLMs. We used the open source \textit{LLaMa-3.1-8B} and \textit{DeepSeek-R1-70B} LLMs as the foundation and used RAG technology to give them domain knowledge. In order to verify whether RAG can effectively improve LLM's expertise in the supply chain field, we tested it using 840 SCMP and 560 CPIM real exam questions. According to the official announcement: SCMP requires a correct rate of 60\% to pass the exam. For the CPIM exam, achieving a 75\% accuracy rate generally ensures a successful outcome. The test results in Section 3.3 show that LLM using RAG has achieved a higher accuracy rate in the professional qualification exam and can reach the level of human supply chain experts (who have passed the professional qualification exam).
\end{itemize}

\paragraph{SCM Scenario Decision}
In order to further verify the professional capabilities of SCM LLM, we constructed a series of supply chain simulation scenarios to evaluate the decision-making capabilities of SCM LLM. This simulation scenario is based on the classical inventory decision problem in SCM and constructs a supply chain with four stages. The decision maker at each stage can be regarded as an independent agent that needs to make a decision in each period ($t = 1, \ldots, T$). Events within each period unfold as follows. First, shipments arrive from upstream suppliers. Second, new orders arrive from downstream customers. In the case of the retailer, the order received at time $t$ is simply retail demand, $D_t$. Third, these new orders are filled, if possible, from current inventory and shipped out. If an order cannot be filled, it is placed in backlog. The period ends with each agent placing an order, $O_{i,t}$, with his upstream supplier. These are the sole decision variables in the game. When making decisions on $O_{i,t}$, each agent not only considers the current orders, inventory, and backlog, but also tries to minimize the chain cost. Chain cost for each period $t$ is defined as the sum of holding and backlog costs across the four stages, expressed as

\begin{equation*}
C(T) = \sum_{i=1}^{4}\sum_{t=1}^{T}\Big[h_i\max\{I_{i,t}, 0\} - s_i\min\{I_{i,t}, 0\}\Big],
\end{equation*}

where $s_i$ and $h_i$ are the unit backlog and inventory holding costs incurred at stage $i$, respectively. And we define inventory as $I_{it}$, where $I_{it} \geq 0$ represents on-hand inventory and $I_{it} < 0$ represents orders in backlog. 

Based on the above framework, we designed a variety of experimental scenarios to evaluate the decision-making capabilities of different agents based on the diversity of market demands. The parameter settings of these scenarios are summarized in Table \ref{tab:scenarios}. Here $\mathscr{U}(a, b)$ denotes the uniform distribution over the interval $[a, b]$; $\mathscr{N}(a, b^2)$ denotes normal distribution with mean $a$ and variance $b^2$; and $\mathscr{C}(a, b)$ denotes a mixed distribution with explanation provided in the next paragraph.
\begin{table}[htbp]
\centering
\renewcommand{\thetable}{\Roman{table}} 
\caption{Parameter settings for different supply chain scenarios.}
\label{tab:scenarios}
\resizebox{\textwidth}{!}{
\begin{tabular}{lccccc}
\hline
Scenario & Constant & Variable & Larger & Seasonal & Normal \\
\hline
Number of Stages & 4 & 4 & 4 & 4 & 4 \\
Number of Periods & 12 & 12 & 12 & 12 & 12 \\
Initial Inventories & [12, 12, 12, 12] & [12, 12, 12, 12] & [12, 12, 12, 12] & [12, 12, 12, 12] & [12, 14, 16, 18] \\
Lead Times & [2, 2, 2, 2] & [2, 2, 2, 2] & [2, 2, 2, 2] & [2, 2, 2, 2] & [1, 2, 3, 4] \\
Demand & 4 & $\mathscr{U}(0, 4)$ & $\mathscr{U}(0, 8)$ & $\mathscr{C}(4, 8)$ & $\mathscr{N}(4, 2^2)$ \\
Product Capacities & [20, 20, 20, 20] & [20, 20, 20, 20] & [20, 20, 20, 20] & [20, 20, 20, 20] & [20, 22, 24, 26] \\
Backlog Costs & [1, 1, 1, 1] & [1, 1, 1, 1] & [1, 1, 1, 1] & [1, 1, 1, 1] & [1, 1, 1, 1] \\
Holding Costs & [0.5, 0.5, 0.5, 0.5] & [0.5, 0.5, 0.5, 0.5] & [0.5, 0.5, 0.5, 0.5] & [0.5, 0.5, 0.5, 0.5] & [0.5, 0.5, 0.5, 0.5] \\
\hline
\end{tabular}
}
\end{table}

These five scenarios correspond to five common market demands. In the first scenario, the market demand is constant at 4 units per period, which is designed to test the basic functions of the model under stable conditions. The second scenario introduces variable market demand, with demand uniformly distributed between 0 and 4 units per period, and adds randomness to evaluate the decision-making ability of the model in the face of fluctuating demand. The third scenario amplifies the market demand changes, with demand uniformly distributed between 0 and 8 units per period, to evaluate the decision-making ability of the model when demand fluctuations become larger. The fourth scenario simulates seasonal demand, dividing the simulation decision process with a total step length of 12 periods into 4 parts, with a step length of 3 cycles in each part. In the first and third parts, demand is uniformly distributed between 0 and 4 units per period; in the second and fourth parts, demand is uniformly distributed between 4 and 8 units per period. This design is designed to evaluate the model's response and decision-making ability in the face of seasonal changes in the market. The fifth scenario faces a normally distributed market demand, with an average demand of 4 units per period and a standard deviation of 2 units. The lead time, initial inventory, and production capacity vary in each stage. This scenario tests the model’s decision-making ability under more realistic demand fluctuations and different operational constraints. Through the above design, we are able to simulate uncertain real markets and thus fully evaluate the model’s decision-making ability.

We let the agents use five different strategies to make decisions in these simulation scenarios, and calculate the chain costs that these strategies can obtain.

\begin{itemize}
    \item \textbf{Base-Stock Policy:} The base inventory strategy is a classic inventory management strategy, under which each stage in the supply chain maintains a constant inventory level, called the base inventory level. This expected inventory level depends on the production capacity of the stage: 
    \begin{equation*}
    \hat{I}_{i,t} = c_i.
    \end{equation*}

    \item \textbf{Tracking Demand Policy:} The demand tracking strategy is an inventory management strategy that relies on historical experience and adjusts orders based on observed demand (or sales) patterns rather than maintaining a constant base inventory level. By dynamically matching supply with actual sales, this strategy ensures the responsiveness and efficiency of the inventory system. In this strategy, the expected inventory is set to:
    \begin{equation*}
    \hat{I}_{i,t} = \bar{S}_{i,t-1}L_i + B_{i,t-1},
    \end{equation*}
    where the average sales in the most recent rounds is
    \begin{equation*}
    \bar{S}_{i,t-1} = \frac{1}{L_{max}} \sum_{\Delta t=1}^{L_{max}} S_{i,t-\Delta t}.
    \end{equation*}
    
    \item \textbf{API:} We call the \textit{DeepSeek-R1-671B} API provided by Deepseek to provide decision suggestions for agents. Because the model has a huge number of parameters and a unique deep learning training design, it can better align with the human thinking process and has great potential in SCM decision-making.

    \item \textbf{AI-Agent (w/o RAG):} The base LLM of \textit{DeepSeek-R1-70B} that we deployed locally. We did not use the RAG system to give it SCM domain knowledge and used it as a baseline for comparison with AI-Agent (w/ RAG).

    \item \textbf{AI-Agent (w/ RAG):} We have deployed a RAG system locally with \textit{DeepSeek-R1-70B} as the base LLM. RAG can give it SCM domain knowledge, and DeepSeek-R1's chain-of-thought can make full use of the knowledge provided by RAG: The model synthesizes insights from multiple documents, referencing relevant passages with clarity, bringing stronger consistency. At the same time, DeepSeek-R1 "thinks aloud" internally, verifying each conclusion through the lens of embedding data, thereby reducing hallucinations to a certain extent. Therefore, AI-Agent (w/ RAG) is likely to show amazing performance in decision-making on professional tasks.
\end{itemize}

Our experimental results (as shown in Table \ref{tab:results} ) highlight the performance of various models in different demand scenarios. The AI-Agent (w/ RAG) model shows very competitive performance, especially compared with AI-Agent (w/o RAG). Although both use the same base LLM, RAG makes the agent's decision-making better in most scenarios. Although sometimes it is not as good as the API model, there is a huge difference in the number of parameters of the LLM they rely on (70B vs 671B). Despite this, the performance of AI-Agent (w/ RAG) is still able to keep up with the API model. This shows the huge potential that vertical LLMs may have in solving tasks in the SCM field.

\begin{table}[htbp]
\centering
\renewcommand{\thetable}{\Roman{table}} 
\caption{Chain cost performance of different strategies under different demand scenarios.}
\label{tab:results}
\begin{tabular}{lccccc}
\hline
Model & Constant & Variable & Larger & Seasonal & Normal \\
\hline
Base-Stock & -296.0 & -523.0 & -392.2 & -274.3 & -322.4 \\
Tracking-Demand & -360.0 & -412.0 & -265.1 & -421.9 & -232.2 \\
API & -187.0 & -340.0 & -226.0 & -439.2 & -172.3 \\
AI-Agent (w/o RAG) & -215.0 & -392.6 & -372.6 & -503.0 & -207.4 \\
AI-Agent (w/ RAG) & -193.0 & -351.3 & -365.2 & -420.6 & -187.6 \\
\hline
\end{tabular}
\end{table}

\section*{Appendix B\quad Exploring Horizontal Games in SCM LLM}

\subsection*{B.1\quad Cournot Model}

\paragraph{Generalized Cournot Equilibrium Model}

In this section, we first present the generalized Cournot equilibrium model. The Cournot model, proposed by French economist Antoine Augustin Cournot in 1838, is a classic framework for analyzing oligopolistic competition. It models the situation where a finite number of firms simultaneously choose output quantities, with the market price determined endogenously by the aggregate supply. This framework is widely used to investigate horizontal competition among firms at the same level in the supply chain, offering foundational insights for equilibrium analysis.

\paragraph{Model Setup}

Assume there are $n$ firms ($i=1,2,\ldots,n$) producing homogeneous products, facing a linear market demand curve:
\[
P(Q) = a - bQ,
\]
where $P$ is the market price, $Q = \sum_{i=1}^n q_i$ is the total output, $q_i$ is the output of firm $i$, and $a, b > 0$ are constants. Let $c_i$ denote the constant marginal cost for firm $i$.

The profit function for firm $i$ is given by:
\[
\pi_i(q_i, q_{-i}) = q_i \left[a - b\sum_{j=1}^n q_j\right] - c_i q_i.
\]

\paragraph{Equilibrium Solution}

The Cournot-Nash equilibrium is characterized by each firm maximizing its profit, given the outputs of the other firms. The first-order condition yields the best response function:
\[
\frac{\partial \pi_i}{\partial q_i} = a - b q_i - b \sum_{j \neq i} q_j - c_i = 0,
\]
\[
\Rightarrow q_i^* = \frac{a - c_i - b \sum_{j \neq i} q_j}{2b}.
\]

\vspace{1em}

\subsubsection*{B.1.1\quad Homogeneous Costs Case}

Assume all firms have identical marginal cost, i.e., $c_i = c$ for all $i$. In the symmetric equilibrium, $q_i^* = q^*$ for all $i$, so:
\[
q^* = \frac{a - c}{b(n+1)}.
\]
The aggregate output is $Q^* = n q^* = \frac{n(a-c)}{b(n+1)}$, and the equilibrium price is $P^* = a - b Q^* = \frac{a + n c}{n+1}$.

\textit{Examples:}
\begin{itemize}
    \item \textbf{Duopoly ($n=2$):}
    \[
    q^* = \frac{a-c}{3b}, \quad Q^* = \frac{2(a-c)}{3b}, \quad P^* = \frac{a+2c}{3}.
    \]
    \item \textbf{Triopoly ($n=3$):}
    \[
    q^* = \frac{a-c}{4b}, \quad Q^* = \frac{3(a-c)}{4b}, \quad P^* = \frac{a+3c}{4}.
    \]
\end{itemize}

\vspace{1em}

\subsubsection*{B.1.2\quad Heterogeneous Costs Case}

The Cournot-Nash equilibrium is obtained by solving this system of $n$ linear equations, one for each firm:
\[
q_i^* = \frac{a - c_i - b\sum_{j \neq i} q_j^*}{2b}, \quad \forall i = 1, \dots, n.
\]
or, equivalently,
\[
2b q_i^* + b \sum_{j \neq i} q_j^* = a - c_i,
\]
\[
b \sum_{j=1}^n q_j^* + b q_i^* = a - c_i.
\]
Stacking these equations for all firms yields a solvable linear system. For $n=3$, the explicit solution is:
\[
\begin{cases}
q_1^* = \dfrac{a - 2c_1 + c_2 + c_3}{4b}, \\[1.5ex]
q_2^* = \dfrac{a - 2c_2 + c_1 + c_3}{4b}, \\[1.5ex]
q_3^* = \dfrac{a - 2c_3 + c_1 + c_2}{4b}.
\end{cases}
\]

In general, for $n$ firms, the equilibrium can be found by solving the following linear system:
\[
\mathbf{A}\mathbf{q}^* = \mathbf{d},
\]
where $\mathbf{A}$ is an $n \times n$ matrix with diagonal elements $2b$ and off-diagonal elements $b$, and $\mathbf{d}$ is a vector with elements $a - c_i$.

The aggregate output and equilibrium market price are given by:
\[
Q^* = \sum_{i=1}^n q_i^* = \frac{n a - \sum_{i=1}^n c_i}{b(n+1)}, 
P^* = \frac{a + \sum_{i=1}^n c_i}{n+1}.
\]

This general formulation allows for the analysis of Cournot equilibria under arbitrary cost heterogeneity.

\subsubsection*{B.1.3\quad Experimental Design}

To further explore the dynamic and adaptive strategic behavior of firms in Cournot competition, we designed and implemented a supply chain horizontal game environment leveraging the DeepSeek-R1 LLM as the decision-making agent for each supplier. The experimental platform simulates repeated Cournot competition among $n$ firms at a given supply chain tier, with well-defined information and interaction protocols.

In each round of the game, all firms (agents) simultaneously choose their production quantities. The market price is then determined by a linear inverse demand function,
\[
P = a - b Q,
\]
where $Q$ is the total market output, and the market capacity is set at 100 units. Each agent observes the previous market outcomes (such as price, total output, and own profit), but does not have access to other agents' current decisions, thus reflecting realistic information asymmetry.

Two experimental scenarios are considered:
\begin{itemize}
    \item \textbf{Duopoly with Homogeneous Costs:} Two suppliers, each with zero marginal production cost ($c_1 = c_2 = 0$), independently decide their output quantities over 10 rounds. Each supplier agent is unaware of the other’s quantity choices in any given round and can only learn and adapt strategies based on observed historical market results.
    \item \textbf{Oligopoly with Heterogeneous Costs:} Three suppliers with asymmetric marginal costs ($c_1 = 15$, $c_2 = 20$, $c_3 = 25$) compete under the same information constraints. The introduction of cost asymmetry allows us to analyze the effect of heterogeneous cost structures on equilibrium outcomes and strategic adaptation.
\end{itemize}

In both scenarios, each supplier agent is instantiated as an independent LLM instance, enabling autonomous reasoning, adaptive learning, and iterative strategic adjustment across rounds. The agents are provided with the market mechanism, their own cost structures, and access only to aggregate market feedback from previous rounds. Through this experimental setup, we are able to systematically investigate how intelligent agents learn and adapt within the horizontal Cournot competition framework, both in symmetric (homogeneous cost) and asymmetric (heterogeneous cost) market environments. The experimental setup leverages two types of messages to guide the LLM agent's behavior: a system message that outlines the market environment and the agent's objectives, and a process message that provides round-specific information and instructions. Below, we present the content of the system and process messages, as well as an example of a single round of agent interaction in the Cournot model. This example demonstrates how the agent receives market information and makes a production decision.

\begin{tcolorbox}[breakable, colback=white, colframe=black!50, sharp corners=south, title=System Message]

You are a rational firm competing in a market with \{\texttt{num\_firms}\} firms. You are \{\texttt{firm\_name}\} (Firm \{\texttt{firm + 1}\}). Your goal is to maximize profit by deciding quantity.

Market mechanism:
\begin{itemize}
    \item Each round you and other firms simultaneously decide your quantities
    \item Market price is determined by the inverse demand function: $P = a - bQ$, where $Q$ is the total quantity of all firms
    \item Your profit equals: Profit = $(P - c) \times q$, where $c$ is your marginal cost, $q$ is your quantity
    \item Other firms are also trying to maximize their profits
    \item Each round you can only observe the previous market results, you cannot know other firms' decisions in advance
\end{itemize}

As a rational entrepreneur, you need to:
\begin{itemize}
    \item Gradually optimize your strategy by observing market reactions
    \item Consider the demand function, your costs, and possible behaviors of other firms
    \item Analyze trends based on historical observations and make the best decision
\end{itemize}
\end{tcolorbox}

\begin{tcolorbox}[breakable, colback=white, colframe=black!50, sharp corners=south, title=Process Message]

It's round \{\texttt{period + 1}\}, you are firm \{\texttt{firm + 1}\} in this market.

Your current state:\\
\{\texttt{get\_state\_description(firm\_state, model\_type)}\}

Market information:\\
\{\texttt{model\_description}\}

You need to decide how many units to produce this round. Consider:
\begin{enumerate}
    \item The market demand function: $P = a - bQ$
    \item Your marginal cost
    \item How your quantity affects market price
    \item How to set quantity to maximize profit: $(P - \text{marginal cost}) \times \text{quantity}$
\end{enumerate}

Please analyze the market situation, then provide your quantity decision.

Start by explaining your reasoning in one or two sentences, then give an integer quantity in square brackets (e.g. [26]).
\end{tcolorbox}

\begin{tcolorbox}[breakable, colback=white, colframe=black!50, sharp corners=south, title=Example of Agent Interaction in the Cournot Model (Heterogeneous Costs)]

\textbf{It's round 1, you are firm 1 in this market.}

\textbf{Your current state:}
\begin{itemize}
    \item Demand function: $P = 100.0 - 1.0 Q$ (where $Q$ is the total quantity of all firms)
    \item Your marginal cost: $15.0$ units/price
\end{itemize}

\textbf{Decision prompt:}\\
You need to decide how many units to produce this round. Consider:
\begin{enumerate}
    \item The market demand function: $P = a - bQ$
    \item Your marginal cost
    \item How your quantity affects market price
    \item How to set quantity to maximize profit: (P $-$ marginal cost) $\times$ quantity
\end{enumerate}

Please analyze the market situation, then provide your quantity decision.\\
Start by explaining your reasoning in one or two sentences, then give an integer quantity in square brackets (e.g. [26]).

\medskip
\hrule
\medskip

\textbf{Low\_cost\_firm Agent (to User Proxy):}

As the low-cost firm, I aim to leverage my cost advantage by producing a higher quantity than competitors, assuming they have higher marginal costs and will produce less. [28]

\medskip
\hrule
\medskip

\textbf{It's round 1, you are firm 2 in this market.}

\textbf{Your current state:}
\begin{itemize}
    \item Demand function: $P = 100.0 - 1.0 Q$ (where $Q$ is the total quantity of all firms)
    \item Your marginal cost: $20.0$ units/price
\end{itemize}

\textbf{Market information:} \textit{(same as above)}

\textbf{Decision prompt:}\\
Please analyze the market situation, then provide your quantity decision.\\
Start by explaining your reasoning in one or two sentences, then give an integer quantity in square brackets (e.g. [26]).

\medskip
\hrule
\medskip

\textbf{Medium\_cost\_firm Agent (to User Proxy):}

Since it's the first round with no prior data, assuming symmetric Cournot competition among 3 firms, the optimal quantity is calculated as $(100 - 20) / (1 \times (3 + 1)) = 20$. [20]

\medskip
\hrule
\medskip

\textbf{It's round 1, you are firm 3 in this market.}

\textbf{Your current state:}
\begin{itemize}
    \item Demand function: $P = 100.0 - 1.0 Q$ (where $Q$ is the total quantity of all firms)
    \item Your marginal cost: $25.0$ units/price
\end{itemize}

\textbf{Market information:} \textit{(same as above)}

\textbf{Decision prompt:}\\
Please analyze the market situation, then provide your quantity decision.\\
Start by explaining your reasoning in one or two sentences, then give an integer quantity in square brackets (e.g. [26]).

\medskip
\hrule
\medskip

\textbf{High\_cost\_firm Agent (to User Proxy):}

As the high-cost firm in a Cournot competition with unknown competitor costs, I'll produce a lower quantity to account for their likely higher production due to lower marginal costs. [16]

\medskip
\hrule
\medskip

\textbf{Market outcome:}
\begin{itemize}
    \item Round = 1
    \item Quantities = \{firm\_0: 28, firm\_1: 20, firm\_2: 16\}
    \item Market price = 36.00
    \item Total quantity = 64
    \item Firms' profits = [588.0, 320.0, 176.0]
    \item Cumulative reward = 1084.00
\end{itemize}

\end{tcolorbox}
\vspace{1em}

\subsection*{B.2\quad Bertrand Model Variants}

\paragraph{Background}

The classical Bertrand model assumes firms compete by setting prices for homogeneous products with identical marginal costs. However, in real-world supply chains, firms often face heterogeneous costs and may also offer differentiated products. Here, we consider two extensions: (1) Bertrand competition with heterogeneous (possibly asymmetric) marginal costs, and (2) Bertrand competition with homogeneous costs but differentiated products.

\vspace{0.5em}

\subsubsection*{B.2.1\quad Bertrand Model with Heterogeneous Marginal Costs}

\paragraph{Model Setup}

Consider $n$ firms, where firm $i$ has constant marginal cost $c_i$. Goods are homogeneous, and consumers always choose the lowest-priced offering. If several firms set the same lowest price, the market is split equally among them.

Let $p_i$ denote the price set by firm $i$. The profit function for firm $i$ is:
\[
\pi_i(p_i, p_{-i}) = 
\begin{cases}
  (p_i - c_i)\, D(p_i), &
    \begin{array}[t]{@{}l@{}}
      \text{if } p_i < \min\limits_{j \neq i} p_j,
    \end{array} \\[2.2ex]
  \dfrac{(p_i - c_i)\, D(p_i)}{k}, &
    \begin{array}[t]{@{}l@{}}
      \text{if } p_i = \min\limits_{j \neq i} p_j\  
      \text{and $k$ firms set the minimum price,}
    \end{array} \\[2.2ex]
  0, &
    \begin{array}[t]{@{}l@{}}
      \text{if } p_i > \min\limits_{j \neq i} p_j,
    \end{array}
\end{cases}
\]
where $D(p_i)$ is the demand at price $p_i$.

\paragraph{Equilibrium Solution}

In the unique pure-strategy Nash equilibrium, the firm with the lowest marginal cost (say, $c_1 = \min\{c_1, c_2, \ldots, c_n\}$) sets its price just below the second-lowest marginal cost:
\[
p_1^* = c_2 - \epsilon,
\]
where $c_2$ is the second-lowest cost among all firms, and $\epsilon$ is an arbitrarily small positive number. All other firms set prices at or above their marginal costs, earning zero market share and profit.

The resulting equilibrium quantity for each firm is:
\[
q_i^* =
\begin{cases}
D(c_2 - \epsilon), & \text{if } c_i = c_1 < c_2, \\[1.5ex]
0, & \text{if } c_i > c_1, \\[1.5ex]
\dfrac{D(c_1)}{k}, & \text{if $k$ firms share the lowest cost $c_1$}.
\end{cases}
\]
That is, the lowest-cost firm serves the entire market at the equilibrium price, and all other firms have zero sales. If $k$ firms share the lowest cost, they equally split the market at that price.

\paragraph{Conclusion}

In the Bertrand model with heterogeneous costs, only the lowest-cost firm serves the entire market, pricing just below the second-lowest cost and earning a positive profit margin. This outcome highlights the competitive advantage of cost leadership in horizontally competitive supply chains.

\vspace{1em}

\subsubsection*{B.2.2 \quad Bertrand Model with Homogeneous Costs and Differentiated Products}

\paragraph{Model Setup}

Assume $n$ firms with identical marginal cost $c$, but products are horizontally differentiated (e.g., through quality, brand, or location). The demand for firm $i$'s product depends on the prices of all firms, i.e., $D_i(p_1, p_2, \ldots, p_n)$. A typical formulation, for example with linear or logit demand, could be:
\[
D_i(p_1, \ldots, p_n) = A - B p_i + C \sum_{j \neq i} p_j,
\]
where $A, B, C > 0$ are parameters reflecting the degree of differentiation and substitution.

Each firm's profit is:
\[
\pi_i(p_1, \ldots, p_n) = (p_i - c) D_i(p_1, \ldots, p_n).
\]

\paragraph{Equilibrium Solution}

The Nash equilibrium is characterized by the set of prices $(p_1^*, \ldots, p_n^*)$ that solve the first-order conditions:
\[
\frac{\partial \pi_i}{\partial p_i} = D_i + (p_i - c) \frac{\partial D_i}{\partial p_i} = 0.
\]
For the linear demand above, this leads to:
\[
A - B p_i + C \sum_{j \neq i} p_j - B(p_i - c) = 0,
\]
\[
\Rightarrow p_i^* = \frac{A + C \sum_{j \neq i} p_j + B c}{2B}.
\]
For symmetric equilibrium ($p_i^* = p^*$ for all $i$), this simplifies to:
\[
p^* = \frac{A + C (n-1)p^* + B c}{2B},
\]
\[
\Rightarrow p^* = \frac{A + B c}{2B - C(n-1)},
\]
provided $2B > C(n-1)$.

Given the symmetric equilibrium price $p^*$, the equilibrium quantity for each firm is:
\[
q^* = D_i(p^*, \ldots, p^*) = A - B p^* + C (n-1) p^* = A + [C(n-1) - B] p^*.
\]
Substituting $p^*$, we obtain:
\[
q^* = A + [C(n-1) - B] \cdot \frac{A + Bc}{2B - C(n-1)} = \frac{A B + B c [C(n-1) - B]}{2B - C(n-1)}.
\]
Thus, each firm's equilibrium price and sales depend on the number of firms, the degree of differentiation, and the cost and demand parameters.

\paragraph{Conclusion}

With product differentiation, the equilibrium price exceeds marginal cost, and all firms can earn positive profits. The degree of differentiation (measured by $C$) determines the intensity of price competition: as products become less substitutable ($C \to 0$), prices approach monopoly levels; as products become more substitutable ($C$ increases), prices approach marginal cost, as in the homogeneous Bertrand case.

\vspace{1em}

\subsubsection*{B.2.3\quad Experimental Design}

To investigate dynamic pricing strategies and equilibrium behaviors in Bertrand competition, we constructed a supply chain horizontal game environment in which LLMs serve as autonomous supplier agents. The market environment simulates repeated price competition between two oligopolistic suppliers across multiple rounds, with well-defined interaction protocols and information constraints.

Market demand is specified by a linear function of the form
\[
Q = a - bP,
\]
where $Q$ denotes total market demand, $P$ is the prevailing market price, and the market size is fixed at 100 units. At the beginning of each round, both supplier agents simultaneously set their respective prices. After prices are submitted, demand is allocated and sales volumes are determined according to the specified market mechanism. Agents are informed of their own sales and profit outcomes, as well as the aggregate market results from the previous round, but do not observe their competitor's current pricing decision.

Two experimental scenarios are examined:
\begin{itemize}
    \item \textbf{Cost Asymmetry:} Two suppliers produce homogeneous products but have different marginal production costs ($c_1 = 20$, $c_2 = 40$). Each supplier independently sets its price in each round over a 10-round horizon, without access to the competitor's real-time decisions. This scenario enables assessment of how cost heterogeneity shapes pricing strategies and market outcomes under otherwise symmetric conditions.
    \item \textbf{Product Heterogeneity:} Both suppliers have identical marginal costs ($c_1 = c_2 = 20$), but the products are differentiated by their degree of substitutability. Product differentiation is parameterized via a heterogeneity coefficient (heterogeneity coefficient $= 0.5$, $1.5$), where higher values indicate less substitutable products and thus more pronounced differentiation. This scenario explores the impact of product heterogeneity on equilibrium pricing and market allocation.
\end{itemize}

Throughout both experiments, all other market parameters and initial conditions are held constant to ensure comparability. Price and sales volume data are collected for each agent over ten rounds, allowing analysis of learning dynamics, adaptive pricing, and the emergence of equilibrium patterns under varying competitive settings. By employing the DeepSeek-R1 LLM as the supplier agent, we are able to observe how advanced AI models respond to changes in cost structure and product differentiation within the classical Bertrand competition framework.

Similar to the experimental design in the Cournot model, the setup utilizes two types of messages to guide the behavior of the LLM agents.  

Below, we provide the exact content of the system message (which sets the overall role and market environment), the process message (which provides round-by-round information and instructions), as well as a sample output from one round of agent interaction.  

This example illustrates how the agent receives information about its environment and makes a price decision in response.

\begin{tcolorbox}[breakable, colback=white, colframe=black!50, sharp corners=south, title=System Message]

You are a rational firm competing in a market with \{\texttt{num\_firms}\} firms. You are \{\texttt{firm\_name}\} (Firm \{\texttt{firm + 1}\}). Your goal is to maximize profit by deciding price.

Market mechanism:
\begin{itemize}
    \item Each round you and other firms simultaneously decide your prices
    \item Your demand function: $q = a - bP + \sum dP'$, where $P'$ are other firms' prices
    \item Parameter $a$ is your potential market size, $b$ is your price elasticity, $d$ is product substitution coefficient
    \item When $d > 0$, rising prices by other firms will increase demand for your product (positive substitution effect)
    \item Your profit equals: Profit $= (P - c) \times q$, where $P$ is your price, $c$ is your marginal cost, $q$ is your sales volume
    \item Other firms are also trying to maximize their profits
    \item Each round you can only observe the previous market results, you cannot know other firms' decisions in advance
\end{itemize}

As a rational entrepreneur, you need to:
\begin{itemize}
    \item Gradually optimize your pricing strategy by observing market reactions
    \item Consider the demand function, product substitutability, your costs, and possible behaviors of other firms
    \item Analyze historical price and sales data to find the optimal price
    \item Think about how price changes affect demand and profit
    \item Remember not to set price below your marginal cost, otherwise you will incur a loss
\end{itemize}
\end{tcolorbox}

\begin{tcolorbox}[breakable, colback=white, colframe=black!50, sharp corners=south, title=Process Message]

It's round \{\texttt{period + 1}\}, you are firm \{\texttt{firm + 1}\} in this market.

Your current state:\\
\{\texttt{get\_state\_description(firm\_state, model\_type)}\}

Market information:\\
\{\texttt{model\_description}\}

You need to decide the price for your product this round. Consider:
\begin{enumerate}
    \item Your demand function: $q = a - bP + \sum dP'$, where $P'$ are other firms' prices
    \item Your marginal cost: \{\texttt{firm\_state['marginal\_cost']}\} units/price
    \item The effect of price on demand volume
    \item Substitution coefficients indicate how other firms' price changes affect your demand
    \item How to set price to maximize profit: (price - marginal cost) $\times$ sales volume
\end{enumerate}

Please analyze the market situation, then provide your price decision.

Start by explaining your reasoning in one or two sentences, then give an integer price in square brackets (e.g. [45]).
\end{tcolorbox}

\begin{tcolorbox}[breakable, colback=white, colframe=black!50, sharp corners=south, title=Example of Agent Interaction in the Bertrand Model (Heterogeneous Costs)]

\textbf{It's round 1, you are firm 1 in this market.}

\textbf{Your current state:}
\begin{itemize}
    \item Demand function parameters: $a = 100.0$, $b = 1.0$
    \item Your marginal cost: $20.0$ units/price
    \item Your demand function: $q = f(\text{all prices})$ based on market demand and product differentiation
    \item Nash equilibrium price for you: $40.00$
    \item Nash equilibrium quantity for you: $60.00$
\end{itemize}

\textbf{Decision prompt:}\\
You need to decide the price for your product this round. Consider:
\begin{enumerate}
    \item Your demand function: $q = a - bP + \sum dP'$, where $P'$ are other firms' prices
    \item Your marginal cost: $20.0$ units/price
    \item The effect of price on demand volume
    \item Substitution coefficients indicate how other firms' price changes affect your demand
    \item How to set price to maximize profit: (price $-$ marginal cost) $\times$ sales volume
\end{enumerate}
Please analyze the market situation, then provide your price decision.\\
Start by explaining your reasoning in one or two sentences, then give an integer price in square brackets (e.g. [45]).

\medskip
\hrule
\medskip

\textbf{Firm\_1 Agent (to User Proxy):}

As the first round with no prior data, setting the Nash equilibrium price maximizes profit assuming competitors act rationally. [40]

\medskip
\hrule
\medskip

\textbf{It's round 1, you are firm 2 in this market.}

\textbf{Your current state:}
\begin{itemize}
    \item Demand function parameters: $a = 100.0$, $b = 1.0$
    \item Your marginal cost: $40.0$ units/price
    \item Your demand function: $q = f(\text{all prices})$ based on market demand and product differentiation
    \item Nash equilibrium price for you: $40.00$
    \item Nash equilibrium quantity for you: $0.00$
\end{itemize}


\textbf{Firm\_2 Agent (to User Proxy):}

In the first round without historical data, setting a price above marginal cost to test demand and profit potential is prudent, considering possible substitution effects if competitors price higher. [70]

\medskip
\hrule

\textbf{Market outcome:}
\begin{itemize}
    \item Round = 1
    \item Prices = \{firm\_0: 40, firm\_1: 70\}
    \item Sales volumes = [60.0, 0.0]
    \item Firms' profits = [1200.0, 0.0]
    \item Cumulative reward = 1200.00
\end{itemize}

\end{tcolorbox}

\section*{Appendix C \quad Exploring Vertical Games in SCM LLM -- Bullwhip Effect and the Beer Game}

\subsection*{C.1\quad Background of the Beer Game}

\begin{figure}[h]
\renewcommand{\thefigure}{\Roman{figure}} 
  \centering
  \includegraphics[width=14cm]{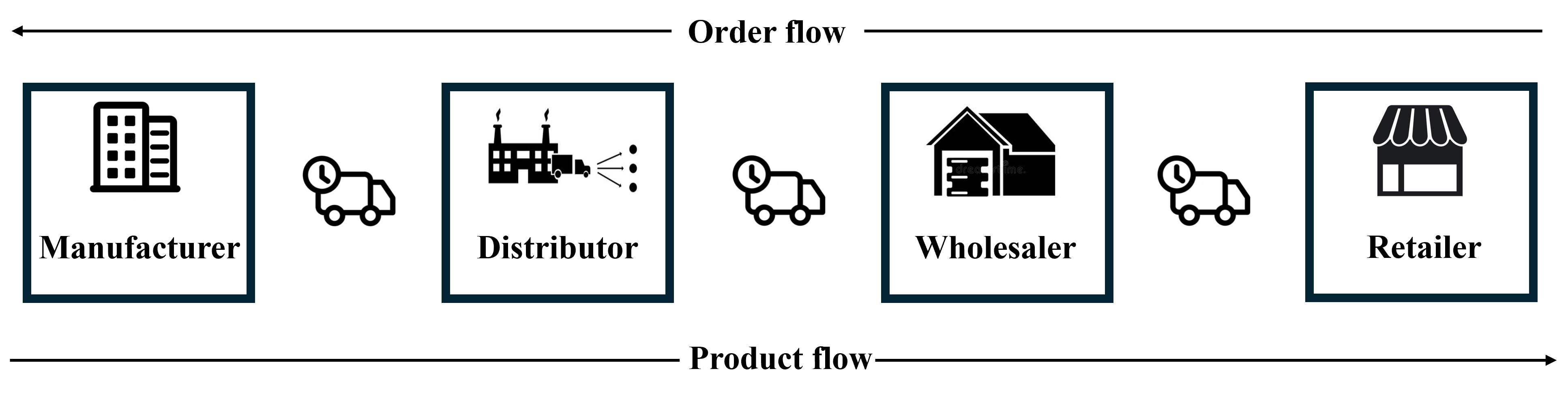}
  \caption{Beer Game.}
  \label{beer game}
\end{figure}

The Beer Game was invented at MIT in the 1960s and has become an iconic experiment in supply chain management (as shown in Figure \ref{beer game}). It simulates the flow of information, goods, and finances among retailer, wholesaler, distributor, and manufacturer roles in a beer supply chain, vividly demonstrating the so-called bullwhip effect -- the amplification of demand variability upstream in the supply chain.

\paragraph{Game Structure and Rules}

\begin{itemize}
    \item \textbf{Participants:} Typically include retailer, wholesaler, distributor, and manufacturer; each may be played by one or more individuals.
    \item \textbf{Objective:} Each tier aims to satisfy downstream orders at the lowest possible inventory and shortage cost.
    \item \textbf{Process:}
    \begin{enumerate}
        \item In each period, the retailer receives customer orders and places orders to the wholesaler; the wholesaler orders from the distributor, who in turn orders from the manufacturer.
        \item Each link in the chain faces information and transportation delays.
        \item Decisions are made based on local information only; participants do not have real-time access to end-customer demand.
    \end{enumerate}
    \item \textbf{Cost Structure:} Both excess inventory and shortages incur costs, and participants must manage a tradeoff between these.
\end{itemize}

\paragraph{Bullwhip Effect}

The bullwhip effect refers to the phenomenon where demand variability is increasingly amplified as information moves upstream along the supply chain, causing upstream firms to face greater uncertainty and volatility. This effect leads to higher inventory costs and reduced service levels across the supply chain.

\subsection*{C.2\quad Experimental Design: Risk Preferences and Information Structures}

Building upon the classical Beer Game, this study systematically investigates the bullwhip effect under three types of risk preferences (risk-averse, risk-neutral, and risk-seeking) and across two information environments: isolated information and information sharing.

\textbf{Experiment design} \quad
We design a classical supply chain experiment comprising four stages: retailer, wholesaler, distributor, and manufacturer. Over 24 periods, the retailer faces market demand drawn from a uniform distribution $U[0,8]$ and places orders upstream; only the retailer knows the demand distribution, while upstream stages make decisions with limited information. Stage lead times are $[2,2,2,3]$, initial inventory is 12 units at each stage, and production capacity is set to 20 units per period per stage. Holding and backlog costs are 0.5 and 1 per unit per period, respectively.

We consider two distinct information structures for the agents:

\begin{itemize}
    \item \textbf{Isolated information:} Each stage observes only its own inventory, backlog, arriving deliveries, and local order history. No additional information is provided beyond local observations.
    \item \textbf{Information sharing:} Each stage has real-time access to the inventory, backlog, and lead times of its immediate upstream partner, enabling more informed ordering decisions. All other experimental parameters remain the same as in the isolated information setting.
\end{itemize}

Agents at each stage are assigned one of three risk preferences:
\begin{itemize}
    \item \textbf{Risk Aversion:} Prioritizes stock safety and over-ordering to avoid stockouts.
    \item \textbf{Risk Neutral:} Balances inventory holding and backlog costs for optimal efficiency.
    \item \textbf{Risk Appetite:} Minimizes inventory while accepting higher risks of stockouts and backlog penalties.
\end{itemize}

\paragraph{Conclusion}

By systematically varying risk preferences and information structures, this experimental framework enables the quantitative analysis of the bullwhip effect and sheds light on how advanced language models adapt their ordering policies under different supply chain environments.

\subsubsection*{C.3\quad Experimental Prompt}

To elicit distinct agent behaviors reflecting different risk preferences and information environments in the beer game supply chain, we design a series of tailored prompts and system messages. These prompts guide the LLM agents to act in accordance with their assigned risk attitudes, and the system message conveys the information available in scenarios with information sharing. The following describes the specific prompt templates and the decision process.

The agent risk preferences are operationalized through explicit textual instructions as shown below.

\begin{tcolorbox}[colback=white, colframe=black!50, sharp corners=south, title=Risk Aversion Agent Prompt]
You are highly risk-averse and prioritize avoiding stockouts at all costs. You should maintain higher inventory levels to ensure you can always meet demand. It's better to have excess inventory than to risk backlog. You should place larger orders earlier to provide a safety buffer.
\end{tcolorbox}

The next prompt guides risk-seeking agent behavior.

\begin{tcolorbox}[colback=white, colframe=black!50, sharp corners=south, title=Risk Appetite Agent Prompt]
You are profit-oriented, and your first goal is to obtain the highest reward. You should keep inventory levels low and place orders in a timely manner. If the loss caused by backlogs affects your reward, you should replenish the stock in time. You should place orders more frequently and adjust your ordering strategy in time to ensure higher rewards.
\end{tcolorbox}

The following prompt describes the risk-neutral tradeoff strategy.

\begin{tcolorbox}[colback=white, colframe=black!50, sharp corners=south, title=Risk Neutral Agent Prompt]
You should balance inventory holding costs with the risk of stockouts. Aim to maintain a moderate inventory level that can handle normal demand fluctuations. Try to balance the costs of backlog with the costs of holding excess inventory.
\end{tcolorbox}

For experiments involving information sharing, agents are provided with an explicit system message clarifying the additional information available.

\begin{tcolorbox}[colback=white, colframe=black!50, sharp corners=south, title=Information Sharing System Prompt]
As part of our information sharing system, you will receive data about upstream stages' Lead Time, Inventory Level, and Current Backlog. Use this information to better anticipate supply chain issues and optimize your ordering decisions.
\end{tcolorbox}

Below are the process messages that provide stage-specific decision guidelines. The manufacturer receives only downstream order information, while the retailer and intermediaries may incorporate upstream visibility in the information sharing scenario.

\begin{tcolorbox}[colback=white, colframe=black!50, sharp corners=south, title=Process Message: Manufacturer (Upstream)]
Guidelines for your decision:
\begin{enumerate}
    \item Consider your current inventory, backlog, and expected future orders.
    \item Account for lead time – you need to place orders in advance.
    \item Analyze patterns in your downstream's ordering history to forecast future demand.
    \item Try to avoid both stockouts and excess inventory.
    \item Open orders should always equal to "expected downstream orders + backlog." If open orders are larger than this, the inventory will rise (once the open orders arrive). If open orders are smaller than this, the backlog will not go down and it may even rise.
\end{enumerate}
Please first explain your reasoning in 1-2 sentences based on the downstream order pattern you observe and your historical performance, then provide your order quantity as a non-negative integer within brackets (e.g. [5]).
\end{tcolorbox}

The following process message is for the retailer and intermediary stages (wholesaler, distributor), which may also reference upstream information in the information sharing scenario.

\begin{tcolorbox}[colback=white, colframe=black!50, sharp corners=south, title=Process Message: Retailer/Intermediaries]
Guidelines for your decision:
\begin{enumerate}
    \item Consider your current inventory, backlog, and expected future orders.
    \item Account for lead time – you need to place orders in advance.
    \item Analyze patterns in your downstream's ordering history to forecast future demand.
    \item Review the information about upstream stages to anticipate potential supply issues.
    \item Try to avoid both stockouts and excess inventory.
    \item Open orders should always equal to "expected downstream orders + backlog." If open orders are larger than this, the inventory will rise (once the open orders arrive). If open orders are smaller than this, the backlog will not go down and it may even rise.
\end{enumerate}
Please first explain your reasoning in 1-2 sentences based on the downstream order pattern you observe, the upstream supply chain information, and your historical performance, then provide your order quantity as a non-negative integer within brackets (e.g. [5]).
\end{tcolorbox}

Finally, we provide a complete example of a single round of agent interaction in the bullwhip effect experiment under the risk-averse setting.

\begin{tcolorbox}[breakable, colback=white, colframe=black!50, sharp corners=south, title={Example of Agent Interaction in the Bullwhip Effect Experiment (Risk-Averse Setting)}]

\textbf{User Proxy (to Retailer Agent):}

Now this is round 7 of 24, and you are at stage 1 (retailer) of 4 in the supply chain.

\textbf{Your current state:}
\begin{itemize}
    \item Lead Time: 2 round(s)
    \item Inventory Level: 8 unit(s)
    \item Current Backlog (you owing to the downstream): 0 unit(s)
    \item Upstream Backlog (your upstream owing to you): 36 unit(s)
    \item Previous Sales (in the recent round(s), from old to new): [4, 6, 2]
    \item Arriving Deliveries (in this and the next round(s), from near to far): [12, 0]
    \item Recent consumer demands (from old to new): [6, 3, 7, 4, 6, 2]
    \item Your market research indicates customer demand varies between 0 and 8 units per round.
    \item Your historical rewards/profits (from old to new): [0, 0, 0, 0, 0, 0]
    \item Your most recent reward was 0, which reflects your reward in the previous round.
    \item Your recent performance has been consistent with your average.
\end{itemize}

\textbf{Decision prompt:} (see guidelines above)

\medskip
\hrule
\medskip

\textbf{Retailer Agent (to User Proxy):}

Given the persistent upstream backlog (36 units) and fluctuating demand up to 8 units, ordering 12 units maintains the safety buffer strategy to preempt stockouts despite unreliable supplier deliveries, ensuring future coverage after 2-round lead time. [12]

\medskip
\hrule
\medskip

\textbf{User Proxy (to Wholesaler Agent):}

Now this is round 7 of 24, and you are at stage 2 (wholesaler) of 4 in the supply chain.

\textbf{Your current state:}
\begin{itemize}
    \item Lead Time: 2 round(s)
    \item Inventory Level: 0 unit(s)
    \item Current Backlog (you owing to the downstream): 36 unit(s)
    \item Upstream Backlog (your upstream owing to you): 160 unit(s)
    \item Previous Sales (in the recent round(s), from old to new): [0, 12, 0]
    \item Arriving Deliveries (in this and the next round(s), from near to far): [0, 0]
    \item Historical orders from your downstream (from old to new): [12, 12, 12, 12, 12, 12]
    \item Your immediate downstream has just placed an order of 12 units.
    \item Your historical rewards/profits (from old to new): [0, -12, -12, -24, -24, -36]
    \item Your most recent reward was -36, which reflects your reward in the previous round.
    \item Your recent decisions have been less profitable than your average performance.
\end{itemize}

\textbf{Decision prompt:} (same as above)

\medskip
\hrule
\medskip

\textbf{Wholesaler Agent (to User Proxy):}

The downstream's consistent 12-unit orders and growing backlog (now 36) necessitate ordering 60 units to cover both backlog and future demand over the 2-round lead time, despite upstream delays. [60]

\medskip
\hrule
\medskip

\textbf{User Proxy (to Distributor Agent):}

Now this is round 7 of 24, and you are at stage 3 (distributor) of 4 in the supply chain.

\textbf{Your current state:}
\begin{itemize}
    \item Lead Time: 2 round(s)
    \item Inventory Level: 0 unit(s)
    \item Current Backlog (you owing to the downstream): 160 unit(s)
    \item Upstream Backlog (your upstream owing to you): 228 unit(s)
    \item Previous Sales (in the recent round(s), from old to new): [0, 0, 8]
    \item Arriving Deliveries (in this and the next round(s), from near to far): [20, 0]
    \item Historical orders from your downstream (from old to new): [12, 36, 36, 36, 36, 36]
    \item Your immediate downstream has just placed an order of 60 units.
    \item Your historical rewards/profits (from old to new): [0, -36, -60, -96, -132, -160]
    \item Your most recent reward was -160, which reflects your reward in the previous round.
    \item Your recent decisions have been less profitable than your average performance.
\end{itemize}

\textbf{Decision prompt:} (same as above)

\medskip
\hrule
\medskip

\textbf{Distributor Agent (to User Proxy):}

The downstream's order jumped to 60, indicating higher demand. To cover the updated backlog (200 after this round) and next two rounds' demand (120), 320 units are needed. With upstream backlog reducing to 208 after receiving 20, order 112 to align with guideline 5 and avoid excess while preventing stockouts. [112]

\medskip
\hrule
\medskip

\textbf{User Proxy (to Manufacturer Agent):}

Now this is round 7 of 24, and you are at stage 4 (manufacturer) of 4 in the supply chain.

\textbf{Your current state:}
\begin{itemize}
    \item Lead Time: 3 round(s)
    \item Inventory Level: 240 unit(s)
    \item Current Backlog (you owing to the downstream): 228 unit(s)
    \item Upstream Backlog (your upstream owing to you): 0 unit(s)
    \item Previous Sales (in the recent round(s), from old to new): [8, 20, 20]
    \item Arriving Deliveries (in this and the next round(s), from near to far): [0, 0, 180]
    \item Historical orders from your downstream (from old to new): [12, 108, 72, 0, 32, 64]
    \item Your immediate downstream has just placed an order of 112 units.
    \item Your historical rewards/profits (from old to new): [0, -108, -180, -172, -184, -228]
    \item Your most recent reward was -228, which reflects your reward in the previous round.
    \item Your recent decisions have been less profitable than your average performance.
\end{itemize}

\textbf{Decision prompt:} (same as above)

\medskip
\hrule
\medskip

\textbf{Manufacturer Agent (to User Proxy):}

The downstream's orders are increasing (32$\rightarrow$64$\rightarrow$112), and given the 3-round lead time, order aggressively to cover future demand and reduce backlog, prioritizing stockout prevention. [256]

\medskip
\hrule
\medskip

\textbf{Market outcome:}
\begin{itemize}
    \item period = 6
    \item action\_dict = \{'stage\_0': 12, 'stage\_1': 60, 'stage\_2': 112, 'stage\_3': 256\}
    \item rewards = \{'stage\_0': 0, 'stage\_1': -48, 'stage\_2': -200, 'stage\_3': -320\}
    \item episode\_reward = -2032
\end{itemize}

\end{tcolorbox}


\end{document}